\UseRawInputEncoding
\documentclass[10pt,final,twocolumn]{IEEEtran}
\long\def\comment#1{}

\usepackage{textcomp}
\usepackage{cite}
\usepackage{graphicx}
\usepackage{subfigure}
\usepackage{psfrag}
\usepackage{url}
\usepackage{stfloats}
\usepackage{amsmath}
\usepackage{amssymb,amsfonts}
\usepackage{amsthm}
\usepackage{float}
\usepackage[usenames]{color}
\usepackage{algorithm,algorithmic}
\usepackage{multirow}
\usepackage{makecell}
\usepackage{threeparttable}

\newtheorem{example}{Example}

\begin{document}

\setlength{\arraycolsep}{0.3em}

\title{A Survey on Distributed Online Optimization and Game
\thanks{}}

\author{Xiuxian Li, Lihua Xie, and Na Li
\thanks{This work was supported by National Natural Science Foundation of China under Grant 62003243 and Grant 62088101, Shanghai Municipal Commission of Science and Technology No. 19511132100, 19511132101, Shanghai Municipal Science and Technology Major Project, No. 2021SHZDZX0100, and Ministry of Education, Singapore, under grant AcRF TIER 1- 2019-T1-001-088 (RG72/19).}
\thanks{X. Li is with the Department of Control Science and Engineering, College of Electronics and Information Engineering, and the Shanghai Research Institute for Intelligent Autonomous Systems, Tongji University, Shanghai, China (Email: xxli@ieee.org).}
\thanks{L. Xie is with the School of Electrical and Electronic Engineering, Nanyang Technological University, 50 Nanyang Avenue, Singapore 639798 (Email: elhxie@ntu.edu.sg).}
\thanks{N. Li is with the Harvard John A. Paulson School of Engineering and Applied Sciences, Harvard University, Cambridge, MA 02138, USA (Email: nali@seas.harvard.edu).}
}

\maketitle

\setcounter{equation}{0}
\setcounter{figure}{0}
\setcounter{table}{0}

\begin{abstract}
Distributed online optimization and game have been increasingly researched in the last decade, mostly motivated by its wide applications in sensor networks, robotics (e.g., distributed target tracking and formation control), smart grids, deep learning, and so forth. In these problems, there is a network of agents who may be cooperative (i.e., distributed online optimization) or noncooperative (i.e., online game) through local information exchanges. And the local cost function of each agent is often time-varying in dynamic and even adversarial environments. At each time, a decision must be made by each agent based on historical information at hand without knowing future information on cost functions. For these problems, a comprehensive survey is still lacking. This paper aims to provide a thorough overview of distributed online optimization and game from the perspective of problem settings, communication, computation, algorithms, and performances. In addition, some potential future directions are also discussed.
\end{abstract}


\section{Introduction}\label{s1}

Optimization (or mathematical programming) and game theory have been extremely popular topics since the last century due to their wide applications across many realms, including computer science, systems and control, finance, biology, medical service, mathematics, machine learning, artificial intelligence, robotics, and so on \cite{osborne1994course,bertsekas1999nonlinear,boyd2004convex}. Optimization and game theory have a similar objective, that is, seeking optimal decision vectors/variables. The difference lies in that for a game, there are usually multiple agents (or players), each aims at computing its own best decision in a noncooperative fashion. It is worth noting that the environments in typical optimization and game theory are often stationary, i.e., the objective/cost/loss functions are time-invariant, which are of limited use in a multitude of practical applications in dynamic environments. For instance, in the target (e.g., robot) tracking problem under either good or atrocious weather condition, the optimal variable can be viewed as the target's position at each time. And the target's position is time-varying as time evolves (note that the target is a part of the environment) \cite{shahrampour2017online}.

\begin{figure}[t]
\centering
\includegraphics[width=2.6in]{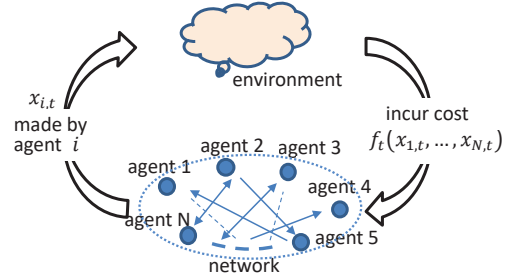}
\caption{Schematic illustration of a general framework for distributed online optimization and game, where $x_{i,t}$ is the decision vector/variable made by agent $i\in[N]$ at time step $t$. The environment is dynamic and even adversarial. $f_t$ is given in a generic form, which can be separable or nonseparable as studied in the literature. And all agents in the network can be cooperative or noncooperative.}
\label{f1}
\end{figure}

In dynamic environments, online learning (OL) for optimization and game (or online optimization/game), as a notable tool for sequential decision making, has become an active research field in recent two decades, mostly because it can well handle a large number of realistic problems in portfolio management, auctions, transportation, smart grids, robotics, dictionary learning, online advertisement placement, online web ranking, neural networks, to just mention a few \cite{zinkevich2003online,mairal2009online,zhou2017incentive}. Online optimization and game usually have three main features: 1) the decision maker does not have access to future information on cost functions in general; 2) the variations of cost functions generally do not obey any statistical distributions; 3) the environments may be even adversarial, i.e., intentionally preventing the decision maker from achieving the best decision (cf. surveys of centralized online optimization/learning \cite{bubeck2011introduction,shalev2012online,hazan2016introduction,mcmahan2017survey}). With the above features, the decision maker has to choose a decision at each time instant only based on historical information at hand, and then the current objective information is revealed. In this setup, it is well known that no algorithms can be leveraged to exactly track the trajectory of best decisions or optimal variables. Usually, two metrics, i.e., regret and competitive ratio, are introduced to measure the performance of proposed algorithms. The metrics basically drive the incurred total cost over a finite horizon to track the lowest cost achieved if knowing all the past and future cost functions in hindsight \cite{zinkevich2003online,sani2014exploiting,yi2021regret}.

With the rapid development of science and technology, as well as the advent of large-scale network and big data in modern life, some limitations of the aforementioned so-called centralized online setup have emerged. For example, no one agent can possess all the information of an optimization/game problem at any time slot, but instead the information is usually distributed over a group of agents who may be geographically dispersed \cite{boyd2011distributed}. In this case, distributed online optimization and game have been put forward. For distributed online optimization (DOO), the global cost function at each time step is unaccessible to any agent, while each of a collection of agents holds partial information on the global cost function and they cooperate to solve the global online optimization by information propagation to their local neighbors. For online game (OG), each agent is often unaware of the cost functions and strategies of other agents at each time. A schematic illustration of the generic framework is presented in Fig. \ref{f1}. Compared with centralized setting, DOO and OG enjoy a plethora of prominent advantages, including privacy preserving, robustness to channel failures, resiliency to cyber-attacks, alleviation of computational burden, etc. \cite{li2021recent}.

Along this line, this paper aims to provide a comprehensive survey on DOO and OG over multi-agent networks by reviewing over one hundred papers published in the last decade, encapsulated primarily from five perspectives: problem settings, communication issues, computation issues, algorithms, and performances. They further include full state information, communication delays, asynchronous algorithms, privacy-preserving, security-guaranteeing, information quantization/compression, full gradient calculation, bandit feedback, projection-free algorithms, etc. To our best knowledge, this paper is the first to report an omnifaceted overview of distributed online optimization and game, which hopefully can motivate and facilitate further study in this field.

{\bf Notations.}
Let $[N]:=\{1,2,\ldots,N\}$ denote an integer set for an integer $N>0$ and $\times_{i=1}^N X_i$ be the Cartesian product of set $X_i$'s. For simplicity, denote by $\mathbf{1}$ and $\mathbf{0}$ column vectors of all entries being $1$ and $0$, respectively, having compatible dimensions from the context. $I_n$ denotes the identity matrix of dimension $n\times n$. Let $\otimes$ be the Kronecker product. Denote by $col(x_1,\ldots,x_N)$ the column vector by piling up vectors $x_i,i\in[N]$ and $x^\top$ the transpose of $x\in\mathbb{R}^n$. Denote by $\langle\cdot,\cdot\rangle$ the standard inner product of two vectors. $\mathcal{B}^n:=\{u\in\mathbb{R}^n|\|u\|\leq 1\}$ and $\mathcal{S}^n:=\{u\in\mathbb{R}^n|\|u\|=1\}$ denote the unit Euclidean ball and sphere of dimension $n>0$, respectively. Denote by ${\bf P}_X(\cdot)$ the projection operator onto a closed convex set $X\subseteq\mathbb{R}^n$ and $[\cdot]_+$ the projection operator onto the nonnegative orthant $\mathbb{R}_+^n$, where $\mathbb{R}_+^n:=\{x=col(x_1,\ldots,x_n)\in\mathbb{R}^n|x_i\geq 0,\forall i\in[n]\}$. Let $\|\cdot\|_*$ denote the dual norm, i.e., $\|x\|_*:=\sup_{\|y\|\leq 1}\langle x,y\rangle$. Denote by $\|\cdot\|$, $\|\cdot\|_1$ and $\|\cdot\|_\infty$ the $\ell_2$-norm, $\ell_1$-norm and $\ell_\infty$-norm, respectively. $\nabla f$ (resp. $\partial f$) represents the gradient (resp. subdifferential) of a function $f$. Let $\mathbb{E}(\cdot)$ and $\mathbb{P}(\cdot)$ be the mathematical expectation and probability, respectively. For two functions $f:\mathbb{R}^n\to\mathbb{R}$ and $g:\mathbb{R}^n\to\mathbb{R}$, $f=O(g)$ denotes that there exists a constant $C>0$ such that $|f(x)|\leq C g(x)$ for all $x\in\mathbb{R}^n$, $f=o(g)$ means $\lim_{x\to\infty}\frac{f(x)}{g(x)}=0$, and $\tilde{O}(\cdot)$ is the same as $O(\cdot)$ up to logarithmic factors.

\section{Problem Settings}\label{s2}

This section aims at encapsulating the studied DOO and OG problems in the literature along with performance metrics and applications. These problems involve a group of $N$ agents, who constitute a multi-agent network where each agent can communicate with its local neighbors via information exchanges. To facilitate the discussion, let us first introduce some fundamentals in graph theory.

\begin{figure}[H]
\centering
\includegraphics[width=2.2in]{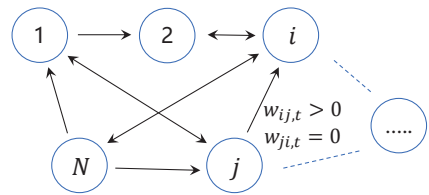}
\caption{Schematic illustration of a directed graph at time $t$, where a directed edge means the availability of information propagation along the directed edge at time $t$.}
\label{f-graph}
\end{figure}

{\bf Graph Theory.} The communication pattern among $N$ agents at each time $t\geq 0$ is captured by a simple graph \cite{west2001}, denoted by $\mathcal{G}_t=(\mathcal{V},\mathcal{E}_t)$, where $\mathcal{V}=[N]$ and $\mathcal{E}_t\subseteq\mathcal{V}\times\mathcal{V}$ are the node (vertex, or agent) and edge sets at time $t$, respectively. The graph can be called {\em communication graph} for all agents. An {\em edge} $(i,j)\in\mathcal{E}_t$ means that agent $i$ can broadcast information to agent $j$ at time $t$. In this case, agent $i$ (resp. $j$) is called an {\em in-neighbor} or simply {\em neighbor} (resp. out-neighbor) of agent $j$ (resp. $i$) at time $t$. Denote by $\mathcal{N}_{i,t}^+:=\{j|(j,i)\in \mathcal{E}_t\}\cup\{i\}$ and $\mathcal{N}_{i,t}^-:=\{j|(i,j)\in \mathcal{E}_t\}\cup\{i\}$ the in-neighbor and out-neighbor sets of agent $i$ at time $t$, respectively. Note that agent $i$ itself is contained in its in-neighbor and out-neighbor sets here, although it may be not the case in graph theory. A graph is called {\em undirected} at time $t$ if and only if $(i,j)\in\mathcal{E}_t$ is equivalent to $(j,i)\in\mathcal{E}_t$ (undirected if it holds for all $t\geq 0$), and {\em directed} otherwise. A {\em directed path} means a sequence of adjacent edges $(i_1,i_2),(i_2,i_3),\ldots,(i_{l-1},i_{l})$. For a stationary graph, i.e., $\mathcal{E}_t$ is fixed, the graph is said to be {\em strongly connected} if any node can be connected to any other node via a directed path. For a time-varying graph $\mathcal{G}_t$, it is called {\em $Q$-strongly connected} for some integer $Q\geq 1$ if any union graph $(\mathcal{V},\cup_{l=0,\ldots,Q-1}\mathcal{E}_{t+l})$ is strongly connected for all $t\geq 0$. Moreover, an information mixing matrix (or adjacency matrix, communication matrix) $W_t=(w_{ij,t})\in\mathbb{R}^{N\times N}$ can be assigned to $\mathcal{G}_t$ such that $w_{ij,t}>0$ when $(j,i)\in \mathcal{E}_t$ and $w_{ij,t}=0$ otherwise. The graph $\mathcal{G}_t$ is called {\em balanced} at time $t$ if $\sum_{j\in\mathcal{N}_{i,t}^+}w_{ij,t}=\sum_{l\in\mathcal{N}_{i,t}^-}w_{li,t}$ for all $i\in[N]$ (called balanced if it holds for all $t\geq 0$), and {\em unbalanced} otherwise. A schematic illustration is provided in Fig. \ref{f-graph}.

To proceed, let us concentrate on explicating the main ingredients of DOO and OG, which typically involves a series of cost/objective/loss functions, including three main settings in the literature, i.e., consensus based DOO, multi-agent coordination based DOO, and OG, as briefly summarized in Table \ref{tb0} and introduced below.

\begin{table*}[ht]
\renewcommand{\arraystretch}{1.3}
\caption{{\upshape Three scenarios of cost functions. Note that other characteristics can generally emerge in all three scenarios. For example,
both fixed and time-varying communication graphs can be considered in all three scenarios, and both stationary or time-varying inequality constraints can appear in all three scenarios. Even in a concrete application, such as distributed estimation in sensor networks, the communication graph can be either stationary or time-varying, which depends upon the specific studied setup.}}
\label{tb0}
\centering
\begin{threeparttable}
\begin{tabular}{c|c|c}
\hline
Scenario & Characteristic & Application examples  \\
\hline\hline
Consensus based DOO & \makecell[c]{Identical decision variable for all agents;\\Cooperative agents} & \makecell[c]{medical diagnosis \cite{mateos2014distributed}; localization in sensor networks \cite{akbari2015distributed};\\ power consumption adjustment for commercial buildings \cite{lesage2020dynamic};\\ distributed estimation in sensor networks \cite{hosseini2013online};\\ distributed target tracking in $2$-D plane \cite{shahrampour2017distributed};\\ multi-class classification \cite{zhang2017projection,jiang2021asynchronous};\\ robot formation control \cite{dixit2019online}} \\
\cline{1-3}
Multi-agent coordination based DOO & \makecell[c]{Agent $i$'s cost relies on its own variable\\ and other agents' variables;\\ Cooperative agents} & \makecell[c]{target surrounding problem for robots in the plane \cite{li2021distributed2};\\ mobile edge computing \cite{wang2021gradient};\\ distributed energy resources for distribution grids \cite{zhou2017incentive}}  \\
\cline{1-3}
Online game & \makecell[c]{Agent $i$'s cost relies on its own variable\\ and other agents' variables;\\ Noncooperative agents} & online Nash-Cournot game \cite{lu2020online,meng2021decentralized,meng2022decentralized}  \\
\hline
\end{tabular}
\end{threeparttable}
\end{table*}

\begin{enumerate}
  \item {\bf Consensus based DOO.} In this setting, the global cost function $f_t$ is in the form
  \begin{align}
  f_t(x)=\sum_{i=1}^N f_{i,t}(x),~~\forall i,j\in[N],       \label{do1}
  \end{align}
  where $x\in\mathcal{X}_{i,t}\subseteq\mathbb{R}^{n}$ with $\mathcal{X}_{i,t}$ being closed and convex in general, and each $f_{i,t}$ is the gradually revealed local cost function of agent $i$ at each time $t\geq 0$. Local cost function means that it is privately revealed to only one agent, but unknown to all other agents.
  \item {\bf Multi-Agent Coordination based DOO.} In this scenario, $f_t$ is of the form
    \begin{align}
  f_t(x_1,\ldots,x_N)=\sum_{i=1}^N f_{i,t}(x_1,\ldots,x_N),                   \label{do2}
  \end{align}
  where $x_i\in\mathcal{X}_{i,t}\subseteq\mathbb{R}^{n_i}$ is the decision variable of agent $i\in[N]$. For notational simplicity, $\mathcal{X}_{i,t}$ is still used here although it may be of different dimension from that in (\ref{do1}). And $f_{i,t}$ is the same as that in (\ref{do1}), except for possibly depending on all agents' variables. Note that it is not necessary for all decision variables $x_i$'s to be identical. An example of this type can be found in Example \ref{ep2} of Section \ref{s2-ss4} later.
  \item {\bf OG.} In this setup, each agent only cares about its own time-varying interest (i.e., time-varying local cost functions). To be specific, different from the cooperation among agents in (\ref{do2}), each agent $i\in[N]$ attempts to minimize its own gradually revealed local cost function $f_{i,t}(x_1,\ldots,x_N)$ subject to $x_i\in\mathcal{X}_{i,t}\subseteq\mathbb{R}^{n_i}$.
\end{enumerate}


{\em Problem Statement.} At each time step $t\geq 0$, each agent $i\in[N]$ makes a decision $x_{i,t}\in\mathcal{X}_{i,t}$ generally based on historical information at hand, i.e., partial or full information on $f_{i,l},\mathcal{X}_{i,l}$ and neighboring information $x_{j,l},j\in\mathcal{N}_{i,l}^+$ for $l\leq t-1$, as well as some current information (e.g., its current position for a robot agent). Then, the environment will reveal some partial or full information on $f_{i,t}$ and $\mathcal{X}_{i,t}$ to agent $i$ along with a suffered loss $f_{i,t}(x_{i,t})$ for consensus based DOO and $f_{i,t}(x_{1,t},\ldots,x_{N,t})$ for multi-agent coordination based DOO and OG. Note that partial information on $f_{i,l}$ here usually means function values at one or several points as studied in one-point or multi-point bandit feedback scenarios \cite{bubeck2011introduction,shalev2012online,hazan2016introduction}, in contrast to complete information on $f_{i,l}$ (e.g., gradients). Meanwhile, partial information sometimes can also mean the neighboring information $x_{j,l},j\in\mathcal{N}_{i,l}^+$ in presence of noises, instead of true information. After that, each agent interacts with its out-neighbors by sharing its some information (e.g., its decision variable at time $t$), and simultaneously receiving neighboring information from its in-neighbors. Then, it continues the above similar process in next time instant until a pre-specified horizon $T>0$ is reached.

Finally, for the aforesaid three scenarios, their corresponding objectives are usually to learn optimal decision strategies for minimizing the following cumulative cost over the total horizon $T$:
\begin{align}
&\min_{x_{l,t}\in\mathcal{X}_{l,t},l\in[N]} \sum_{t=1}^T \sum_{i=1}^N f_{i,t}(x_{i,t}),                     \label{2}\\
&\min_{x_{l,t}\in\mathcal{X}_{l,t},l\in[N]} \sum_{t=1}^T \sum_{i=1}^N f_{i,t}(x_{1,t},\ldots,x_{N,t}),       \label{2b}\\
&\min_{x_{i,t}\in\mathcal{X}_{i,t}} \sum_{t=1}^T f_{i,t}(x_{i,t},x_{-i,t}),                        \label{do3}
\end{align}
where, for brevity, $x_{-i,t}$ denotes the variables of all agents except $i$, i.e., $x_{-i,t}:=col(x_{1,t},\ldots,x_{i-1,t},x_{i+1,t},\ldots,x_{N,t})$. Note that (\ref{do3}) for OG is made for individual agent $i$, instead of the total cost as in the former two scenarios, since each agent only cares about its own interest in OG. Along this line, recent works \cite{lu2020online,meng2021decentralized,meng2022decentralized} have studied online game with/without constraints.



\subsection{Further Discussion on Problem Settings}

Although three main categories of problems have been presented above, to summarize various studied problems in more details, we further classify these problems from other viewpoints, as discussed below.

{\bf Cost Functions.} From the perspective of cost functions, DOO problems can be classified as the conventional case, i.e., a single local cost $f_{i,t}$ for each agent $i\in[N]$ at time $t$, which is the most widely investigated in the literature \cite{hosseini2013online,yan2012distributed,mateos2014distributed,koppel2015saddle}, and other cases, as follows.

\begin{itemize}
  \item {\em Composite Functions.} In general, the cost function $f_t$ can be smooth or nonsmooth. In order to exploit the fine structure of $f_t$ when it is nonsmooth, the cost function can be considered as a sum of two functions, one is smooth and the other is a nonsmooth regularizer, i.e., composite optimization. Such problems can be naturally found in realistic applications (e.g., linear regression) involving low-rank, sparsity, monotonicity, and so forth \cite{dixit2020online,yuan2020distributed}.
  \item {\em Nonseparable Global Objectives.} Some practical applications, such as resource allocation \cite{raginsky2011decentralized,lee2016coordinate}, may encounter the scenario where global $f_t(x_1,\ldots,x_N)$ is nonseparable, but each agent $i$ only maintains a part (or a coordinate) $x_i$ of the whole variable $col(x_1,\ldots,x_N)$.
\end{itemize}

{\bf Constraints.} The constraint set $\mathcal{X}_{i,t},i\in[N]$ is in a general form, meaning that $\mathcal{X}_{i,t}$ can be the entire Euclidean space, i.e., $\mathcal{X}_{i,t}=\mathbb{R}^n$ (or $\mathbb{R}^{n_i}$), or can be of the following form:
\begin{align}
\mathcal{X}_{i,t}=\{x_i\in \mathcal{X}_i|~\text{s.t. (in)equality constraints}\},            \label{3}
\end{align}
where $\mathcal{X}_i\subseteq\mathbb{R}^n$ (or $\mathbb{R}^{n_i}$) is a closed convex set, standing for the simple set constraint for agent $i$, on which it is often relatively efficient to perform the projection operation. Note that $\mathcal{X}_i$ may be also time-varying, but it has been considered to be time-invariant in most of existing literature, thus written as a fixed one here. More details are introduced in the following.

\begin{itemize}
  \item {\em Simple Set Constraints.} In this case, the set $\mathcal{X}_{i,t}$ for each $i\in[N]$ is simply $\mathcal{X}_i$ (or a common set $\mathcal{X}$), that is, $\mathcal{X}_{i,t}=\mathcal{X}_i$ (or $\mathcal{X}_{i,t}=\mathcal{X}$) for all $i\in[N],t\in[T]$, where $\mathcal{X}_i$ or $\mathcal{X}$ is usually assumed to be compact, although having some exceptions, e.g., \cite{mateos2014distributed,zhang2019distributed,han2021differentially}. This case has been widely investigated in the literature \cite{hosseini2013online,yan2012distributed,koppel2015saddle,nedic2015decentralized}. It is noteworthy that the compactness of $\mathcal{X}_i$ or $\mathcal{X}$ is frequently postulated even in centralized online optimization \cite{zinkevich2003online}. However, this condition can be eliminated in some cases, such as the unconstrained case with central functions \cite{mateos2014distributed} or Lipschitz continuous gradients \cite{zhang2019distributed,han2021differentially}.
  \item {\em Local Equality/Inequality Constraints.} This case has been specially studied, mostly due to the facts that constrained sets sometimes consist of equality and/or inequality constraints either on which it is computationally prohibitive to perform the projection operation or which have particular structures that can be elegantly exploited when developing algorithms (e.g., equality constraints of the form $Ax+By=c$ for variables $x\in\mathbb{R}^{n_1},y\in\mathbb{R}^{n_2}$ with $A\in\mathbb{R}^{m\times n_1},B\in\mathbb{R}^{m\times n_2},c\in\mathbb{R}^m$, often handled by the alternating direction method of multipliers (ADMM) \cite{akbari2018individual}) \cite{yuan2017adaptive,sharma2020distributed,yuan2020distributed2,yuan2021distributed}.
  \item {\em Coupled Equality/Inequality Constraints.} In this case, constraints are coupled (or separable) across all the agents. And each agent can only access partial information on constraint functions at each time \cite{li2020distributed,yi2020distributed,yi2020distributed2,li2020onlineLE}, that is,
      \begin{align}
      \mathcal{X}_{t}&=\{x\in \mathcal{X} | \sum_{i=1}^N h_{i,t}(x_i)=\mathbf{0}_l\},~~~            \label{5}\\
      \mathcal{X}_{t}&=\{x\in\mathcal{X} | \sum_{i=1}^N g_{i,t}(x_i)\leq \mathbf{0}_m\},        \label{6}
      \end{align}
      where $h_{i,t}:\mathbb{R}^{n_i}\to\mathbb{R}^l$ and $g_{i,t}:\mathbb{R}^{n_i}\to\mathbb{R}^m$, as constraint functions at time $t$, are gradually revealed only to agent $i$. And $\mathcal{X}_t$ simply denotes $\mathcal{X}_{i,t}$ for each $i\in[N]$, representing a global set constraint, which constitutes an identical constraint set imposed by all agents in this case. Moreover, it is worthwhile to note that all agents' variables are not required to be nonidentical in (\ref{5}) and (\ref{6}), and the separable constraints in (\ref{5}) and (\ref{6}) can simultaneously appear in a single problem.
  \item {\em Control Systems.} It is easy to observe that no system dynamics is considered in the aforesaid problems. In physical world, agents involved in real applications are often subject to some physical dynamics, such as bicycle dynamics for robots and Euler-Lagrange dynamics for manipulators. Therefore, to better suit these applications, control system dynamics has been integrated into DOO recently \cite{chang2021regret,yu2022continuous}, which can be viewed as a sort of constraint for DOO and OG.
\end{itemize}

{\bf Time-Varying Scenarios.} According to distinct scenarios, the models of time-varying cost and/or constraint functions studied in the literature can be summarized as follows, for which the oblivious model is the most widely studied one.
\begin{itemize}
  \item {\em Oblivious Model.} In this case, all cost/constraint functions are determined at the beginning by the environment or adversary, which cannot be changed in the process of learning.
  \item {\em Stochastic Model.} This model means the existence of some stochastic components for local cost functions, incurred by sampling data or other factors. Usually, the local cost function $f_{i,t}$ is dependent on an additional random variable $\xi_{i,t}$, which is subject to a generally unknown distribution $D_{i,t}$, i.e., $f_{i,t}(\cdot;\xi_{i,t})$ at time $t$ \cite{zhao2019decentralized}.
\end{itemize}

{\bf Function Properties.}
\begin{itemize}
  \item {\em Convex/Nonconvex Functions.} In existing literature, the convex case has been more frequently studied until now, where cost functions and constraint functions (if available) are all convex \cite{hosseini2013online,yan2012distributed,mateos2014distributed,li2020distributed}. Meanwhile, the nonconvex case has been less addressed, where either cost or constraint functions are nonconvex, usually leading to more complex problems than the convex ones. To date, only a few papers considered the nonconvex case with only simple set constraints \cite{lu2019online,lu2021online,jiang2022distributed}.
  \item {\em Strong Convexity and/or Smoothness.} To derive better performance, several elegant function properties can be particularly leveraged, such as strong convexity and smoothness, which are also realistic in many applications (e.g., linear regression with a quadratic regularizer). For example, strong convexity and smoothness have been harnessed to improve the performance for DOO in \cite{zhang2019distributed}.
\end{itemize}

\subsection{Performance Metrics}\label{s2.1}

With the above problem settings, how to evaluate whether a developed algorithm is good or not is apparently an important task for effective algorithm proposals. For online learning, there are generally three classes of metrics which have been leveraged in the literature, as summarized below. For simplicity, the following metrics are introduced for the consensus based DOO due to its most popularity, but similar metrics can be also defined for (\ref{2b}) and (\ref{do3}).
\begin{enumerate}
  \item {\bf Dynamic Regret.} Equipped with the objective in (\ref{2}), one direct metric is to compare the accumulated cost against a sequence of any comparators $\{\varrho_t\}_{t=1}^T$, satisfying $\varrho_t\in\mathcal{X}_t:=\cap_{i=1}^N \mathcal{X}_{i,t}$. Then, the {\em dynamic regret} is defined as \cite{zinkevich2003online}
      \begin{align}
      D\text{-}Reg(\{\varrho_t\}):=\sum_{t=1}^T \sum_{i=1}^N f_{i,t}(x_{i,t})-\sum_{t=1}^T f_t(\varrho_t).       \label{7}
      \end{align}
      For (\ref{7}), there are two paramount special cases: 1) the case by setting $\varrho_t=x_t^*:=\mathop{\arg\min}_{x\in\mathcal{X}_t}f_t(x)$, sometimes also called {\em restricted dynamic regret}; and 2) the case by setting $\varrho_t=x^*:=\mathop{\arg\min}_{x\in\cap_{t=1}^T\mathcal{X}_t}\sum_{t=1}^T f_t(x)$, which is usually called {\em static regret}, denoted specially as $S\text{-}Reg$, i.e.,
      \begin{align}
      S\text{-}Reg:=\sum_{t=1}^T \sum_{i=1}^N f_{i,t}(x_{i,t})-\sum_{t=1}^T f_t(x^*),       \label{8}
      \end{align}
      where an underlying assumption is $\cap_{t=1}^T\mathcal{X}_t\neq \emptyset$. It is easy to see that the static regret is to compare the total cost with the minimal cost with respect to a decision variable which is the same one (i.e., time-invariant) all over the time horizon. It should be noted that the dynamic regret is generally not sublinear with respect to $T$ in the worst case, hence requiring some regularities, such as path variation/length and gradient variation, on the comparator sequence or the cost function sequence \cite{jadbabaie2015online,yang2016tracking}.

      Another metric related to dynamic regret is so-called {\em individual dynamic regret} (or {\em local/agent dynamic regret}) for each $j\in[N]$ (e.g., \cite{mateos2014distributed,lesage2020dynamic}), defined as
      \begin{align}
      D\text{-}Reg_j(\{\varrho_t\}):=\sum_{t=1}^T f_{t}(x_{j,t})-\sum_{t=1}^T f_t(\varrho_t).       \label{9}
      \end{align}
      That is, the compared cumulative cost is with respect to the decision variables of agent $j$, instead of all agents' variables, which can measure the special performance for individual agent in some sense. Similarly, one can define {\em individual/local/agent static regret} (or simply {\em individual/local/agent regret}) (e.g., \cite{koppel2015saddle,akbari2018individual}) as
      \begin{align}
      S\text{-}Reg_j:=\sum_{t=1}^T f_{t}(x_{j,t})-\sum_{t=1}^T f_t(x^*),~\forall j\in[N].       \label{10}
      \end{align}
      Note that associated with individual regrets in (\ref{9}) and (\ref{10}), the regrets in (\ref{7}) and (\ref{8}) can be called {\em network regrets}.
  \item {\bf Adaptive Regret.} The dynamic regret is defined over the entire horizon $t\in[T]$, i.e., from a global perspective. Another metric, called {\em adaptive regret}, is proposed from a local perspective. That is, it aims at comparing the cumulative cost over a subset of the entire horizon $[T]$, which includes two versions, i.e., strongly and weakly adaptive regrets. For {\em strongly adaptive regret} \cite{daniely2015strongly}, the purpose is to minimize the maximal static regret over a fixed time interval, say $\tau>0$, which is defined as
      \begin{align}
      SA\text{-}Reg(\tau)&:=\max_{[r,r+\tau-1]\subseteq [T]}\sum_{t=r}^{r+\tau-1} \sum_{i=1}^N f_{i,t}(x_{i,t})     \nonumber\\
      &\hspace{0.5cm}-\min_{x\in\cap_{t=1}^T\mathcal{X}_t}\sum_{t=r}^{r+\tau-1} f_t(x).       \label{11}
      \end{align}
      Furthermore, the {\em weakly adaptive regret} \cite{hazan2007adaptive} aims to minimize the maximal static regret over any contiguous time intervals, specifically defined by
      \begin{align}
      WA\text{-}Reg&:=\max_{[r,s]\subseteq [T]}\sum_{t=r}^s \sum_{i=1}^N f_{i,t}(x_{i,t})     \nonumber\\
      &\hspace{0.5cm}-\min_{x\in\cap_{t=1}^T\mathcal{X}_t}\sum_{t=r}^s f_t(x).       \label{12}
      \end{align}
      It is easy to observe that weakly adaptive regret is greater than strongly adaptive regret, which is dependent on the time interval $\tau$ and will incur different SA-Reg($\tau$) values for distinct $\tau\in[1,T]$.
  \item {\bf Competitive Ratio.} Besides the above performance metrics, another natural metric is {\em competitive ratio} (CR for brevity) \cite{shi2018competitive}. It is defined by the division between the cumulative cost and the minimal overall cost over time horizon $T$, i.e.,
      \begin{align}
      CR:=\frac{\sum_{t=1}^T \sum_{i=1}^N f_{i,t}(x_{i,t})}{\sum_{t=1}^T f_t(x_t^*)},       \label{13}
      \end{align}
      where $x_t^*=\mathop{\arg\min}_{x\in\mathcal{X}_t}f_t(x)$. In this case, it is generally assumed that $\sum_{t=1}^T f_t(x_t^*)>0$.

      Additionally, another type of metric related to comparative ratio is also employed in some literature. That is, an online algorithm is said to have a competitive ratio $c_r\geq 1$ \cite{krumke2002online,eghbali2016designing} if there exists a constant $c_r'\geq 0$, which is independent of $T$, such that
      \begin{align}
      \sum_{t=1}^T \sum_{i=1}^N f_{i,t}(x_{i,t})\leq c_r\sum_{t=1}^T f_t(x_t^*)+c_r',       \label{14}
      \end{align}
      which is only different from (\ref{13}) by a constant $c_r'$.
\end{enumerate}

{\em ``Good'' Performance:} With the above performance measures, a proposed algorithm is usually declared ``good'' if dynamic regret or adaptive regret is bounded by a sublinear term in $T$ (i.e., $o(T)$) and other problem-dependent regularities, or the competitive ratio is a constant, i.e., $CR\leq c_r$ for some constant $c_r>0$ or satisfying (\ref{14}) with an additional constant $c_r'>0$.

It is worth noting that all the above metrics are introduced in the context of deterministic setting, which however can be easily accommodated to suit the stochastic scenario by replacing the cumulative cost with one in mathematical expectation with respect to any randomness in the context, e.g., stochastic cost functions, stochastic gradients, sample datasets, and stochastic algorithms, etc. Additionally, the aforesaid metrics can be also applied to the nonconvex setup in the sense of local optimum, and other special metrics for the nonconvex case were also proposed in the literature, such as the regret defined by first-order optimality condition \cite{lu2021online}.

It is also noteworthy that besides the aforementioned metrics, another metric is to directly track the time-varying optimal variables, i.e., analyzing $\|x_{i,t}-x_t^*\|$ with $x_t^*$ being the optimal decision variable at time $t$. It is often bounded by a constant that is determined by problem parameters (e.g., Lipschitz constant, gradient's bound, etc.) and network factors (e.g., agents' number, network connectivity, etc.). This line of problems is often called (centralized/distributed) {\em time-varying optimization} \cite{simonetto2018dual,cao2019dynamic,huang2019distributed,bastianello2021distributed,ospina2021time}. Similarly, when the problem is to consider the fixed point computing for a series of operators in real Hilbert space, its goal is to track the sequence of time-varying fixed points, often called {\em time-varying operator} \cite{dall2019convergence,bastianello2021stochastic}. This type of metric provides another perspective, but is different from the aforementioned metrics in DOO and OG, for which a recent survey can be found in \cite{andrea2020time}. Also note that time-varying operator is now only addressed in the centralized manner. Nonetheless, the time-invariant scenario has been taken into account in the distributed fashion in recent years \cite{fullmer2018distributed,li2020distributed2,li2021distributed3,li2020dotdop}, leaving the time-varying case as one of future directions.

Among the aforementioned performance metrics, the most widely exploited ones in online learning are static and dynamic regrets, followed by competitive ratio (more frequently leveraged in metrical task system (MTS) \cite{borodin1992optimal}, etc.) and then adaptive regrets in the literature. In DOO and OG, static and (restricted) dynamic regrets are mainly employed in the literature, and actually, it does not have a clear rule to guide which metrics to choose. Roughly speaking, both regrets can be selected for DOO and OG, but a good performance on static regret cannot imply a good performance on dynamic regret in general. Therefore, dynamic regret can be generally considered for DOO and OG, which is more general, including static and restricted dynamic regret as two special cases. However, dynamic regret is generally more difficult to analyze than static regret. Generally speaking, they provide different metrics for measuring designed online algorithms in online learning from diverse perspectives, which have vague relationships as discussed in the sequel. Up to date, quantitative relationship between dynamic regret and adaptive regret remains unclear and an attempt for simultaneously minimizing both dynamic and adaptive regrets was made in recent work \cite{zhang2020minimizing}. Moreover, static regret is usually incompatible with competitive ratio, that is, they often cannot perform well simultaneously \cite{andrew2013tale}. However, in recent work \cite{daniely2019competitive} the authors proposed a method for simultaneously ensuring a small static regret and a guaranteed constant competitive ratio for a special $N$-experts $D$-switching cost problem. Nonetheless, the problem of concurrently achieving the best static/dynamic regret and competitive ratio even in a general centralized convex setup remains open.

\subsection{Connection to Related Problems}

Multi-agent reinforcement learning (MARL) is for a family of agents to learn to make their best decisions or actions by interacting with an environment, in order to maximize their own total expected discounted return \cite{nowe2012game}. MARL is usually based on Markov decision processes which are employed to describe the possible state transition under a previous state and committed actions of all agents. Therefore, DOO and OG, as formulated at the beginning of Section \ref{s2}, are essentially different from MARL.

Concept drift means that the relationship between the input data and the target variable (mainly in online supervised learning, e.g., regression or classification tasks) changes over time, in order to distinguish the conventional scenario where the data are drawn from an unknown but fixed probability distribution. In concept drift, both posterior and evidence distributions can be time-varying, including real drift and virtual drift, and its objective is to achieve high accuracy for learning tasks \cite{ditzler2015learning,gama2014survey}. In contrast, DOO and OG are differently formulated with substantial distinct performance metrics, where no specific tasks are considered.

Online federated learning (OFL) is a popular paradigm for resolving some challenges encountered in modern machine learning, such as streaming data \cite{dai2022addressing}. As in federated learning, there are a family of edge nodes/devices with a central/global server. At each round, the global server broadcasts its latest global model (i.e., the parameter of a global function) to all edge nodes, and each node then updates its local model by the received global model and its local real-time streaming data. Then, all nodes send their updated models to the global server for some sort of dynamic aggregation. The main difference lies in that OFL has a global server to aggregate the information of all nodes, while no such global coordinator exists in DOO and OG which are formulated in a distributed manner.

Distributed machine learning (DML) is a popular paradigm for processing a substantial amount of data in machine learning by distributing the workload across a collection of machines \cite{verbraeken2020survey,peteiro2013survey}. In DML, the studied topologies usually have special structures, such as centralized, tree-like topology, parameter-server topology, peer-to-peer network. However, the topology in DOO can be arbitrary, although often assumed connected or $Q$-strongly connected (cf. graph theory for the notions in Section \ref{s2}). In DML, the methods for a model improvement include evolutionary algorithms (or genetic algorithms), stochastic gradient descent based algorithms, rule-based machine learning algorithms, etc. Some algorithms (e.g., stochastic gradient descent) can aid in designing online algorithms in DOO. However, DML basically studies stationary problems (i.e., fixed dada distributions), while DOO and OG consider time-varying problems where cost/constraint functions can be arbitrarily changing, thus requiring different performance metrics as well.

Distributed optimization, also called decentralized optimization in some works, is aimed for a group of agents to collaboratively tackle a global optimization problem, where the global cost function is composed of a sum of local cost functions \cite{nedic2018network,xin2020general,yang2019survey}. And each local cost function is only privately accessible to each individual agent. The kind of problem can find numerous applications, such as robotics and power systems \cite{molzahn2017survey}, and is also closely related to signal processing \cite{vlaski2022networked} (e.g., distributed learning in wireless sensor networks \cite{predd2006distributed}). Distributed optimization is similar to DOO, but the differences lie in that cost functions in DOO are arbitrarily changing, instead of stationary as in distributed optimization, and they have different performance metrics. However, the methods of algorithm design in distributed optimization provide significant insights and contributions to the algorithm design for DOO, although the analysis of DOO is intrinsically distinct from that in distributed optimization. For instance, the distributed (sub)gradient descent algorithm in distributed optimization is also employed to handle DOO, e.g., distributed autonomous online learning (DAOL) \cite{yan2012distributed} (cf. Algorithm \ref{ag1} later).

Game theory is a popular paradigm for decision making, where each agent selfishly minimizes its own cost \cite{shoham2008multiagent,nowe2012game}. It should be noted that cost functions involved in classical game theory are often fixed, i.e., time-invariant. In contrast, the main feature in OG is that cost functions of each agent are time-varying and even adversarial.

In summary, as generic mathematical frameworks, DOO and OG are not formulated for specific learning tasks. In DOO and OG, cost functions can arbitrarily change in general in dynamic and even adversarial environments, thus resulting in that solution to past functions cannot help solve new ones. Moreover, DOO and OG focus on developing mathematical algorithms to minimize some performance metrics, such as dynamic regret, as discussed in Section \ref{s2}-B.

\subsection{Applications}\label{s2-ss4}

Numerous applications of DOO and OG can be found in reality, including the binary classification problem from medical diagnosis \cite{mateos2014distributed}, collaborative localization in sensor networks \cite{akbari2015distributed}, demand and response in commercial buildings \cite{lesage2020dynamic}, distributed dynamic sparse recovery problem \cite{dixit2020online}, distributed estimation in sensor networks \cite{hosseini2013online}, distributed target tracking in $2$-D plane \cite{shahrampour2017distributed}, target surrounding problem for robots in the plane \cite{li2021distributed2}, mobile edge computing \cite{wang2021gradient}, multi-class classification \cite{zhang2017projection,jiang2021asynchronous}, regularized linear regression \cite{yuan2020distributed}, robot formation control \cite{dixit2019online}, support vector machines (SVM) for binary classification \cite{koppel2015saddle}, distributed energy resources for distribution grids \cite{zhou2017incentive}, and so on. As examples, three applications in the literature are briefly introduced below.

\begin{example}[Distributed Target Tracking \cite{shahrampour2017distributed}]
Consider a slowly moving (nearly constant velocity) target in a plane with independently evolving horizontal and vertical components of its position. The state of the target is composed of four components at each time instant, i.e., horizontal position, vertical position, horizontal velocity, and vertical velocity. As a result, the state at each time $t$, denoted by $x_t^*\in\mathbb{R}^4$, obeys the physical motion dynamics as follows:
\begin{align}
x_{t+1}^*=Ax_t^*+\nu_t,        \label{ap1}
\end{align}
where $\nu_t$ is the system noise, and
\begin{align}
A=\left(
    \begin{array}{cc}
      1 & \delta \\
      0 & 1 \\
    \end{array}
  \right)\otimes I_2        \label{ap2}
\end{align}
with $\delta>0$ being the sampling interval. The objective is to track $x_t^*$ of the target by a sensor network of $N$ agents in a collaborative manner.

The agents are located on an $M\times M$ grid for some integer $M>0$, aiming at tracking the moving target cooperatively (i.e., learning the target position and tracking it). At each time step $t$, an observation $z_{i,t}\in\mathbb{R}$ on partial information $x_{t}^*$ (i.e., one of its noisy coordinates) can be performed by each agent $i$, satisfying
\begin{align}
z_{i,t}=e_{k_i}^\top x_t^*+\omega_{i,t},       \label{ap3}
\end{align}
where $e_{k}$ is the $k$th unit vector in $\mathbb{R}^4$, i.e., all entries are zero with the $k$th entry being $1$, and $\omega_{i,t}\in\mathbb{R}$ is the observation noise. $k_i$ can be randomly chosen or pre-specified for each agent in order to ensure that every component of $x_t^*$ is observed by at least one sensor. That is, agent $i$ knows $z_{i,t}$, $k_i$, and $x_{j,l}$'s for $j\in\mathcal{N}_i^+,l\leq t-1$ at time $t$. In this scenario, each agent cannot observe the target on its own, but $x_t^*$ is globally identifiable from the viewpoint of the entire network. Then the local square cost function is of the form
\begin{align}
f_{i,t}(x)=(z_{i,t}-e_{k_i}^\top x)^2,       \label{ap4}
\end{align}
and the global cost function is
\begin{align}
f_t(x)=\sum_{i=1}^N f_{i,t}(x),       \label{ap5}
\end{align}
which boils down to the consensus based DOO. After making an estimation $x_{i,t}$ by agent $i$ at time $t$, an instantaneous cost $f_{i,t}(x_{i,t})$ is incurred. And its goal can be written as Eq. (\ref{2}).
\end{example}

\begin{example}[Distributed Target Surrounding \cite{li2021distributed2,carnevale2021distributed}]\label{ep2}
Consider a moving robot target in $\mathbb{R}^2$ space, denoted by $x_0(t)$, which tends to be attacked by a group of $M>0$ intruders. Meanwhile, there are a family of $N$ robots, which aim at protecting the target by surrounding it. And each defender is only aware of its nearby intruders, that is, no single defender can grab the information on all intruders. However, by combining all defenders' information, the attack from all intruders can be avoided in general. For simplicity, let us consider the case of $M=N$, and agent $i\in[N]$ is only aware of the intruder $i\in[N]$. Note that to better protect the target, it is preferable to drive all defenders to some positions such that the target is located at the center of all defenders. Therefore, the loss should be proportional to the distance from the average position of all defenders to the target. That is, the local loss function is of the form
\begin{align}
f_{i,t}(x_{i,t},\nu(x_t)),~~~\nu(x_t):=\frac{1}{N}\sum_{i=1}^N x_{i,t},      \label{exam2-1}
\end{align}
and the global loss function is
\begin{align}
f_t(x_1,\ldots,x_N)=\sum_{i=1}^N f_{i,t}(x_{i,t},\nu(x_t)),      \label{exam2-2}
\end{align}
which is an instance of multi-agent coordination based DOO (\ref{do2}). In this problem, at each time $t>0$, each agent $i\in[N]$ makes a decision by deciding to move to a position $x_{i,t}$ (possibly subject to a position constraint $x_{i,t}\in\mathcal{X}_{i,t}$) based on known information of intruder $i$'s position at time $t$ and received in-neighboring positions $x_{j,l}, j\in\mathcal{N}_{i,l}^+,l\leq t-1$, and then a loss $f_{i,t}(x_{i,t})$ is revealed to agent $i$. As a result, its objective is exactly in the form of (\ref{2b}) but with $f_{i,t}$ being in the specific form (\ref{exam2-1}).
\end{example}

\begin{example}[Online Nash-Cournot Game \cite{lu2020online,meng2021decentralized}]\label{em3}
The online Nash-Cournot game is denoted by $\Gamma([N],\mathcal{X},f_t)$, where $\mathcal{X}=\mathcal{X}_{1}\times\cdots\times\mathcal{X}_{N}$ denotes the action set of $N$ firms (or players), representing production constraints and market capacity constraints. And $f_t=(f_{1,t},\ldots,f_{N,t})$ is the cost function with $f_{i,t}$ being the local cost function of firm $i$ at time slot $t$. Note that the same production is produced by all firms, which interact with each other via some underlying communication graph. Denote by $x_i\in\mathbb{R}$ the production quantity of firm $i$. In view of some instable factors such as the production cost, marginal costs and the demand price, the firm $i$'s production cost and demand price are given as $p_{i,t}(x_i)=\alpha_i(t)x_i$ and $d_{i,t}(x)=\beta_i(t)-\sum_{i=1}^N x_i$ for some $\alpha_i(t),\beta_i(t)>0$, respectively. As a result, the local cost function of firm $i$ is of the form
\begin{align}
f_{i,t}(x_i,x_{-i})=p_{i,t}(x_i)-x_i d_{i,t}(x),~~~\forall t\in[0,T]         \label{exam3}
\end{align}
where $x_{-i}=col(x_1,\ldots,x_{i-1},x_{i+1},\ldots,x_N)$. At time $t$, firm $i$ knows $p_{i,s}$ $(s\leq t)$, $x_{i,l}$ and $d_{i,l}$ ($l\leq t-1$), but without the knowledge of $d_{i,t}$. The goal of firm $i$ is to learn a good strategy (i.e., its production quantity) $x_{i,t}$ to minimize its cost $f_{i,t}$ at each time $t$. Once $x_{i,t}$'s are made by all firms, the incurred cost $f_{i,t}(x_{i,t},x_{-i,t})$ is revealed to firm $i$ for $i\in[N]$. In this case, its aim is to minimize its total cost over $T$ rounds, amounting to (\ref{do3}) in the setup of OG.
\end{example}

\begin{example}[Distributed Kernel-based Online Regression \cite{liang2022kernel}]
This problem is adapted from \cite{liang2022kernel}. There are $N$ agents and $N$ streaming data $({\bf x}_{i,t},y_{i,t})$ with inputs ${\bf x}_{i,t}\in\Omega\subseteq\mathbb{R}^d$ and labels $y_{i,t}\in\mathbb{R}$. Each agent aims to collaboratively learn a global model $F({\bf x}):\Omega\to\mathbb{R}$ to predict the label of an unknown incoming data ${\bf x}$. The local estimator $F_{i,t}$ of $F$ by agent $i$ at time $t$ is of the form
\begin{align}
F_{i,t}({\bf x})=\sum_{s=1}^{t}\xi_{i,s} Ker_i({\bf x}_{i,s},{\bf x}),~~~{\bf x}\in\Omega,       \label{exam4-1}
\end{align}
where $\xi_{i,s}$'s are real coefficients, related to the prediction error $e_{i,s}:=F_{i,s-1}({\bf x}_{i,s})-y_{i,s}$, and $Ker_i$ is the kernel function of agent $i$ (e.g., Gaussian kernel $\exp(-\kappa_i\|{\bf x}_{i,s}-{\bf x}\|^2)$ with some $\kappa_i>0$). When new data sample ${\bf x}_{i,t}$ arrives at time $t$, the label prediction $F_{i,t-1}({\bf x}_{i,t})$ can be made by agent $i$. Then, the true label $y_{i,t}$ is revealed with an instantaneous incurred loss
\begin{align}
f_{i,t}(F_{i,t-1})=\ell(F_{i,t-1}({\bf x}_{i,t})-y_{i,t}),       \label{exam4-2}
\end{align}
where $\ell$ can be any loss functions, e.g., the quadratic loss function and hinge loss \cite{liang2022kernel}. The overall goal can be written as (\ref{2}), falling into the setup of consensus based DOO.
\end{example}

\section{State-of-the-art Algorithms}

This section aims to introduce existing efficient algorithms for resolving DOO and OG. In view of vast algorithms in the literature, only the algorithms with state-of-the-art performances are presented, whose performances will be elaborated in Section \ref{sec6}. For simplicity, let us denote by $\partial f$ any one subgradient in this section, although $\partial f$ represents the differential for a nondifferentiable function $f$, which amounts to the gradient $\nabla f$ when $f$ is differentiable.

To proceed, let us first introduce a few basic classical algorithms for handling stationary optimization which are helpful for understanding the state-of-the-art algorithms introduced later.

{\em Projected Gradient Descent.} For solving the problem
\begin{align}
\min_{x\in \Omega}~f(x)        \label{opt}
\end{align}
with $\Omega\subseteq\mathbb{R}^n$ being closed and convex, the standard projected gradient descent is given as
\begin{align}
x_{t+1}={\bf P}_\Omega\big(x_{t}-\alpha_t\nabla f_{t}(x_{t})\big),     \label{PGD}
\end{align}
where $\alpha_t>0$ is the stepsize or learning rate \cite{boyd2004convex}.

{\em Mirror Descent.} To introduce this algorithm, an important notion is the Bregman divergence, which is defined, with respect to a differentiable and strictly/strongly convex function $\phi(x)$, as
\begin{align}
\mathcal{D}_\phi(x,y):=\phi(x)-\phi(y)-\nabla \langle \phi(y),x-y\rangle.
\end{align}
As a general distance-measuring function, the Bregman divergence includes the traditional Euclidean norm and the Kullback-Leibler divergence as two special cases \cite{boyd2004convex}. Then, the standard mirror descent algorithm for solving (\ref{opt}) is of the form
\begin{align}
x_{t+1}=\mathop{\arg\min}_{x\in\Omega}\{\alpha_t \langle x,\nabla f(x_{t})\rangle+\mathcal{D}_\phi(x,x_{t})\},   \label{MD}
\end{align}
where $\alpha_t>0$ is the stepsize \cite{boyd2004convex}. Note that projected gradient descent can be regarded as a special case of the mirror descent algorithm (\ref{MD}) by choosing $\phi(x)=\|x\|^2/2$.

{\em Primal-Dual Algorithm.} Consider the following problem
\begin{align}
\min_{x\in\Omega}~f(x),~~s.t.~~g(x)\leq {\bf 0}_m,       \label{gopt}
\end{align}
where $f:\mathbb{R}^n\to\mathbb{R}$ and $g:\mathbb{R}^n\to\mathbb{R}^m$ are cost and constraint functions, respectively. Note that affine equality constraints can be also considered in (\ref{gopt}), which are omitted here. To compute the optimal variable, the standard primal-dual algorithm \cite{boyd2004convex} is of the form
\begin{align}
x_{t+1}&={\bf P}_\Omega\big(x_{t}-\alpha_t\nabla_x L(x_t,\lambda_t)\big),     \nonumber\\
\lambda_{t+1}&=\big[\lambda_t+\alpha_t\nabla_\lambda L(x_t,\lambda_t)\big]_+,       \label{pda}
\end{align}
where $\alpha_t>0$ is the stepsize, $\lambda_t\in\mathbb{R}_+^m$ is the dual variable, $\nabla_x L(x,\lambda)$ and $\nabla_\lambda L(x,\lambda)$ denote gradients of $L$ with respect to $x$ and $\lambda$, respectively, and $L$ is the Lagrangian function (or some augmented Lagrangian) defined as
\begin{align}
L(x,\lambda):=f(x)+\lambda^\top g(x).
\end{align}

It is worth noting that gradients in the above algorithms can be replaced with subgradients when dealing with nondifferentiable functions. In addition, distributed extensions of the above algorithms can be found in \cite{nedic2018network,xin2020general,yang2019survey} for solving distributed optimization. In what follows, different state-of-the-art algorithms for tackling DOO and OG are introduced according to distinct scenarios, ranging from the simple case to intricate cases.

\subsection{Consensus based DOO with only Simple Set Constraint}

In this case, the common set constraint $x\in\mathcal{X}\subseteq\mathbb{R}^n$ is imposed for consensus based DOO, for which three interesting setups are considered in the following, i.e., full information feedback, one-point bandit feedback, and projection-free algorithms.

For the case with full information feedback (i.e., functions are observed with full/true (sub)gradients), the typical algorithms are based on either (sub)gradient descent or mirror descent, as detailed in Algorithm \ref{ag1} \cite{yan2012distributed} (studying static regret), Algorithm \ref{ag2} \cite{shahrampour2017distributed} (studying dynamic regret for convex cost functions), and Algorithm \ref{ag3} \cite{eshraghi2021improving} (studying dynamic regret for strongly convex cost functions). Wherein, the term $\sum_{j\in\mathcal{N}_i^+} w_{ij}x_{j,t}$ aims to mix the neighboring information in order to eventually achieve a common variable for all agents. And $K_t$ round communications among neighboring agents are required for decision variables and gradients at each time $t$ in Algorithm \ref{ag3}.

\begin{algorithm}
 \caption{Distributed Autonomous Online Learning (DAOL) (Full information feedback; Static regret) \cite{yan2012distributed}}     \label{ag1}
 \begin{algorithmic}[1]
  \STATE \textbf{Input:} Initial points $x_{i,0},i\in[N]$; double stochastic mixing matrix $W=(w_{ij})\in\mathbb{R}^{N\times N}$; and maximum iterations $T$.
  \STATE \textbf{Iterations:} Step $t\geq 0$, update for each $i\in[N]$:\\
\hspace{0.5cm}Agent $i$ receives information $x_{j,t}$ from its in-neighbors;
\begin{align}
x_{i,t+1}={\bf P}_\mathcal{X}\Big(\sum_{j\in\mathcal{N}_i^+} w_{ij}x_{j,t}+\alpha_t\partial f_{i,t}(x_{i,t})\Big);              \label{A1}
\end{align}
\hspace{0.5cm}Send $x_{i,t+1}$ to its out-neighbors.\\
In (\ref{A1}), $\alpha_t>0$ is the stepsize or learning rate, designed as $\alpha_t=\frac{1}{2\sqrt{t}}$ for convex $f_{i,t}$'s and $\alpha_t=\frac{1}{2\mu t}$ for $\mu$-strongly convex $f_{i,t}$'s.
 \end{algorithmic}
\end{algorithm}

\begin{algorithm}
 \caption{Decentralized Online Mirror Descent (DOMD) (Full information feedback; Dynamic regret) \cite{shahrampour2017distributed}}     \label{ag2}
 \begin{algorithmic}[1]
  \STATE \textbf{Input:} Initial points $x_{i,0},y_{i,0},i\in[N]$; double stochastic mixing matrix $W=(w_{ij})\in\mathbb{R}^{N\times N}$; and maximum iterations $T$.
  \STATE \textbf{Iterations:} Step $t\geq 0$, update for each $i\in[N]$:
\begin{align*}
\hat{x}_{i,t+1}&=\mathop{\arg\min}_{x\in\mathcal{X}}\{\alpha \langle x,\nabla f_{i,t}(x_{i,t})\rangle+\mathcal{D}_\phi(x,y_{i,t})\},     \\
x_{i,t+1}&=A\hat{x}_{i,t+1},        \\
y_{i,t+1}&=\sum_{j\in\mathcal{N}_i^+}w_{ij}x_{j,t+1},
\end{align*}
where $\alpha>0$ is the stepsize or learning rate, designed as $\alpha=\sqrt{(1-\sigma_2(W))(C_T+2R^2)/T}$, $\sigma_2(W)$ denotes the second largest singular value of $W$, $R^2:=\sup_{x,y\in\mathcal{X}}\mathcal{D}_\phi(x,y)$, $C_T:=\sum_{t=1}^T\|x_{t+1}^*-A x_t^*\|$ with $x_t^*=\mathop{\arg\min_{x\in\mathcal{X}}}f_t(x)$. And $A$ is a common knowledge on the deviation of the minimizer sequence, i.e., $x_{t+1}^*=Ax_t^*+e_t$ with $e_t$ being an unknown and unstructured noise.
 \end{algorithmic}
\end{algorithm}

\begin{algorithm}
 \caption{DOMD-MADGC (Full information feedback; Dynamic regret) \cite{eshraghi2021improving}}     \label{ag3}
 \begin{algorithmic}[1]
  \STATE \textbf{Input:} Initial points $x_{i,0}\in\mathcal{X},i\in[N]$; stepsize $\alpha$; double stochastic mixing matrix $W_t$; time horizon $T$.
  \STATE \textbf{Iterations:} Step $t\geq 0$, update for each $i\in[N]$:\\
\hspace{1.2cm} Set $K_t=\lceil \frac{-2\log t}{\log\sigma_2(W_t)}\rceil$
\begin{align*}
y_{i,t}&=\sum_{j=1}^N((W_t)^{K_t})_{ij}x_{j,t}     \\
\hat{g}_{i,t}&=\sum_{j=1}^N((W_t)^{K_t})_{ij}\nabla f_{j,t}(x_{j,t})   \\
x_{i,t+1}&=\mathop{\arg\min}_{x\in\mathcal{X}}\big\{\langle x,\hat{g}_{i,t}\rangle+\frac{1}{\alpha}\mathcal{D}_\phi(x,y_{i,t})\big\}
\end{align*}
where $\lceil\cdot\rceil$ represents the ceiling function, $((W_t)^{K_t})_{ij}$ means the $(i,j)$th entry of matrix $(W_t)^{K_t}$, and the stepsize $\alpha$ satisfies $\frac{l_\phi-\mu_\phi}{\mu}<\alpha<\frac{l_\phi}{\mu}$. Note that $\mu,\mu_\phi,l_\phi$ are the constants of strong convexity of $f_{i,t},\phi$ and Lipschitzness of $\nabla \phi$, respectively.
 \end{algorithmic}
\end{algorithm}

For the case with one-point bandit feedback, where the only information revealed to each agent $i\in[N]$ is the function value at the committed decision, instead of full subgradients, the state-of-the-art algorithm is a special case of Algorithm \ref{ag5} provided later by removing inequality constraints, thus omitted here.

Regarding projection-free algorithms, whose goal is to cope with computationally heavy projections onto a complicated set via locally light computations, the state-of-the-art algorithm is listed in Algorithm \ref{ag4}. Wherein, when $f_{i,t}$'s are convex, i.e., $\mu=0$, main parameters are chosen as
\begin{align}
K=M=\sqrt{T},~~\beta=\frac{N^{1/4}T^{3/4}L}{\sqrt{1-\sigma_2(W)}R},    \label{ag4-p1}
\end{align}
where $\sigma_2(W)$ denotes the second largest singular value of $W$, $L$ is the Lipschitz constant of $f_{i,t}$, and $R>0$ is a constant such that $\mathcal{X}\subseteq R\mathcal{B}^n$. Meanwhile, when $f_{i,t}$'s are strongly convex, i.e., $\mu>0$, main parameters are selected as
\begin{align}
K=M=T^{2/3}(\ln T)^{-2/3},~~\beta=\mu K.    \label{ag4-p2}
\end{align}

\begin{algorithm}
 \caption{Distributed Block Online Conditional Gradient (D-BOCG) (Static regret) \cite{wan2021projection}}     \label{ag4}
 \begin{algorithmic}[1]
  \STATE \textbf{Input:} Feasible set $\mathcal{X}, x_{in}\in\mathcal{X}, \beta, M, K$, and the strong convexity constant $\mu\geq 0$.
  \STATE \textbf{Initialization:} Choose $x_{i,0}=x_{in},z_{i,0}={\bf 0}_n$ for $i\in[N]$.
\STATE {\bf for $m=1,\ldots,T/K$ do}
\STATE ~~{\bf for each local agent $i\in[N]$ do}\\
\hspace{0.3cm} define $F_{i,m}(x)=z_{i,m}^\top x+\frac{(m-1)\mu K}{2}\|x\|^2+\beta\|x-x_{in}\|^2$\\
\hspace{0.3cm} $\hat{g}_{i,m}={\bf 0}_n$\\
\STATE \hspace{0.4cm}{\bf for $t=(m-1)K+1,\ldots,mK$ do}\\
\hspace{0.4cm} play $x_{i,m}$ and observe $\nabla f_{i,t}(x_{i,m})$\\
\hspace{0.4cm} $\hat{g}_{i,m}=\hat{g}_{i,m}+\nabla f_{i,t}(x_{i,m})$\\
\STATE \hspace{0.3cm} {\bf end for}\\
\hspace{0.5cm} $x_{i,m+1}=CG(\mathcal{X},M,F_{i,m}(x),x_{i,m})$\\
\hspace{4.0cm}//refer to Algorithm \ref{ag4-1} for CG\\
\hspace{0.5cm} $z_{i,m+1}=\sum_{j\in\mathcal{N}_i^+}w_{ij}z_{j,m}+\hat{g}_{i,m}-\mu K x_{i,m}$
\STATE \hspace{0.25cm}{\bf end for}
\STATE {\bf end for}
 \end{algorithmic}
\end{algorithm}

\begin{algorithm}
 \caption{Conditional Gradient (CG, also known as Frank-Wolf) \cite{wan2021projection}}     \label{ag4-1}
 \begin{algorithmic}[1]
  \STATE \textbf{Input:} Feasible set $\mathcal{X}$, $M$, $F(x)$, and $x_{in}$
  \STATE $c_0=x_{in}$
\STATE {\bf for $\tau=0,1,\ldots,M-1$ do}\\
\hspace{0.3cm} $v_\tau\in \mathop{\arg\min}_{x\in \mathcal{X}}\nabla F(c_\tau)^\top x$\\
\hspace{0.3cm} $s_\tau=\mathop{\arg\min}_{s\in[0,1]}F(c_\tau+s(v_\tau-c_\tau))$\\
\hspace{0.3cm} $c_{\tau+1}=c_\tau+s_\tau(v_\tau-c_\tau)$\\
\STATE {\bf end for}
\STATE {\bf return $x_{out}=c_M$}
 \end{algorithmic}
\end{algorithm}

In Algorithm \ref{ag4}, a surrogate loss function $F_{i,m}$ of $\sum_{k=1}^t f_{i,k}$ is constructed for each agent $i\in[N]$ at time $t$, and the variable $\hat{g}_{i,m}$ accumulates the gradient $\nabla f_{i,t}$ over $K$ rounds to reduce the gradient communication complexity. Meanwhile, by making use of $F_{i,m}$ and $\hat{g}_{i,m}$, the classical conditional gradient algorithm (or Frank-Wolf algorithm) is leveraged by running $M$ rounds in each block $m=1,\ldots,T/K$.

Note that Algorithm \ref{ag4} is also extended to handle the one-point bandit feedback case in \cite{wan2021projection}, where the static regret is taken into consideration.

\subsection{DOO with Inequality Constraints}

In this case, the cost functions can be either consensus based or multi-agent coordination based, and meanwhile, inequality constraints can be either uncoupled or coupled across all the agents in a network.

As for the case with global time-invariant inequality constraints, i.e., the decision must satisfy $c_s(x)\leq 0$ for $s=1,\ldots,p$, where each $c_s:\mathbb{R}^n\to\mathbb{R}$ is convex and differentiable with bounded gradients. Note that bounded gradients can be usually satisfied under the condition of constrained sets' compactness. The following only considers the one-point bandit feedback case, which, however, can be easily accommodated to deal with the full information feedback case by replacing true gradients with gradient estimators. For the scenario with one-point bandit feedback, one of important ideas is to leverage the following one-point gradient estimator:
\begin{align}
\tilde{\nabla}f_{i,t}(x_{i,t}):=\frac{n}{\epsilon_t}f_{i,t}(x_{i,t}+\epsilon_t u_{i,t})u_{i,t},        \label{ope}
\end{align}
where $u_{i,t}$ is randomly and uniformly selected on the unit sphere and $\epsilon_t>0$ is sufficiently small. Then, an algorithm, called DOCO-LTC (meaning distributed online convex optimization with long-term constraints), is proposed in \cite{yuan2021distributed} as given in Algorithm \ref{ag5}, where parameters are chosen as
\begin{align}
\alpha_t=\frac{1}{apB_g^2T^\kappa},~\eta_t=\frac{1}{T^\kappa},~\epsilon_t=\frac{1}{T^b},~\vartheta=\frac{1}{R_{\mathcal{X}}T^b}.  \label{ope1}
\end{align}
In (\ref{ope1}), $B_g>0$ is an upper bound on (sub)gradients of $f_{i,t}$'s and $c_s$'s, $a>1$ is a constant, $\kappa\in(0,1)$, $b=\kappa/3$, and $R_{\mathcal{X}}>0$ is a constant such that $\|x\|\leq R_\mathcal{X}$ for all $x\in\mathcal{X}$. Algorithm \ref{ag5} is basically a variant of primal-dual algorithm, where an augmented Lagrangian $\tilde{L}_{i,t}(x,\lambda)$ is employed for agent $i$ at time $t$. It is worth noting that the inequality constraint $c_s(x)\leq 0$ is equivalently transformed to $[c_s(x)]_+\leq 0$ with the purpose of studying a stricter performance on inequality constraints, i.e., $\sum_{t=1}^T \sum_{s=1}^p[c_s(x_{i,t})]_+$, instead of the conventional $\big[\sum_{t=1}^T \sum_{s=1}^p c_s(x_{i,t})\big]_+$.

\begin{algorithm}
 \caption{DOCO-LTC (One-point bandit feedback; Static regret) \cite{yuan2021distributed}}     \label{ag5}
 \begin{algorithmic}[1]
  \STATE \textbf{Input:} Double stochastic mixing matrix $W_t=(w_{ij,t})\in\mathbb{R}^{N\times N}$; stepsize $\alpha_t$; regularization parameter $\eta_t$; exploration parameter $\epsilon_t$; and shrinkage parameter $\vartheta\in(0,1)$.
  \STATE {\bf Initialization:} $x_{i,0}={\bf 0}_n$ and $\lambda_{i,0}={\bf 0}_p$ for $i\in[N]$.
  \STATE \textbf{Iterations:} Step $t\in[0,T]$, update for each $i\in[N]$:
\begin{align*}
y_{i,t}&=x_{i,t}-\alpha_t[\tilde{\nabla}f_{i,t}(x_{i,t})+\sum_{s=1}^p[\lambda_{i,t}]_s\partial[c_s(x_{i,t})]_+],        \\
x_{i,t+1}&={\bf P}_{(1-\vartheta)\mathcal{B}}\Big(\sum_{j=1}^N w_{ij,t}y_{j,t}\Big),                                     \\
\lambda_{i,t+1}&=\mathop{\arg\max}_{\lambda\in\mathbb{R}_+^p}\tilde{L}_{i,t}(x_{i,t+1},\lambda),
\end{align*}
where $\tilde{L}_{i,t}(x,\lambda):=\tilde{f}_{i,t}(x;\epsilon)+\sum_{s=1}^p[\lambda]_s[c_s(x)]_+-\frac{\eta_t}{2}\|\lambda\|^2$, $\tilde{f}_{i,t}(x;\epsilon):=\mathbb{E}_v[f_{i,t}(x+\epsilon v)]$ is the smoothed loss function, $v$ is a vector uniformly distributed over the unit ball, $[\lambda]_s$ denotes the $i$th component of the vector $\lambda\in\mathbb{R}^p$, and $\mathcal{B}:=\{x\in\mathbb{R}^n|\|x\|\leq R_\mathcal{X}\}$ is a ball containing $\mathcal{X}$ for some $R_{\mathcal{X}}>0$.
 \end{algorithmic}
\end{algorithm}

Consider the scenario with local time-varying inequality constraints, that is, $g_{i,t}(x)\leq {\bf 0}$ is the local inequality constraint for agent $i$ at time $t$, where $g_{i,t}:\mathcal{X}\to\mathbb{R}^{m_i}$ is a sequence of local constraint functions. For this case, a recent efficient algorithm is given in Algorithm \ref{ag6} \cite{yi2021regret2}, which is also extended to handle the two-point bandit feedback case in \cite{yi2021regret2}. In Algorithm \ref{ag6}, similar to Algorithm \ref{ag5}, the inequality constraint $g_{i,t}\leq {\bf 0}$ is equivalently replaced with $[g_{i,t}]_+\leq {\bf 0}$ for addressing a stricter constraint performance, where $[\cdot]_+$ is performed componentwise. In the meantime, when updating the dual variable $\lambda_{i,t}$, the function $[g_{i,t}(x)]_+$ is approximated as $[g_{i,t}(x_{i,t})]_+ +(\partial[g_{i,t}(x_{i,t})]_+)^\top(x-x_{i,t})$.

\begin{algorithm}
 \caption{Distributed Online Algorithm (DOA) (Full information feedback; Static and dynamic regrets) \cite{yi2021regret2}}     \label{ag6}
 \begin{algorithmic}[1]
  \STATE \textbf{Input:} Non-increasing and positive $\alpha_t, \beta_t$ and $\gamma_t$; double stochastic mixing matrix $W_t=(w_{ij,t})\in\mathbb{R}^{N\times N}$.
  \STATE {\bf Initialization:} $x_{i,0}\in\mathcal{X}$ and $\lambda_{i,0}={\bf 0}_{m_i}$ for $i\in[N]$.
  \STATE \textbf{Iterations:} Step $t\in[0,T]$, update for each $i\in[N]$:
\begin{align*}
z_{i,t+1}&=\sum_{j=1}^N w_{ij,t}x_{j,t},       \\
\varpi_{i,t+1}&=\partial f_{i,t}(x_{i,t})+\partial[g_{i,t}(x_{i,t})]_+\lambda_{i,t},    \\
x_{i,t+1}&= {\bf P}_{\mathcal{X}}(z_{i,t+1}-\alpha_{t+1}\varpi_{i,t+1}),       \\
\lambda_{i,t+1}&=\big[(1-\beta_{t+1}\gamma_{t+1})\lambda_{i,t}+\gamma_{t+1}([g_{i,t}(x_{i,t})]_+    \\
&\hspace{0.4cm}+(\partial[g_{i,t}(x_{i,t})]_+)^\top(x_{i,t+1}-x_{i,t}))\big]_+,
\end{align*}
where parameters are set as $\alpha_t=\alpha_0/t^\kappa$, $\beta_t=1/t^\kappa$, and $\gamma_t=1/t^{1-\kappa}$ with $\alpha_0>0$ and $\kappa\in(0,1)$.
 \end{algorithmic}
\end{algorithm}

Regarding time-varying coupled inequality constraints, i.e.,
\begin{align}
\sum_{i=1}^N g_{i,t}(x_i)\leq {\bf 0}       \label{ag7-e1}
\end{align}
with $x_i\in\mathbb{R}^{n_i}$ being the local decision variable of agent $i$, where $g_{i,t}:\mathcal{X}\to\mathbb{R}^m$ is the constraint function and gradually revealed only to agent $i$ for all $i\in[N]$. The state-of-the-art algorithms are provided in Algorithm \ref{ag7} \cite{li2020onlineLE} (studying static regret for convex cost and constraint functions), Algorithm \ref{ag8} \cite{yi2020distributed} (studying static regret for strongly convex cost functions and dynamic regret for convex cost and constraint functions), and Algorithm \ref{ag9} \cite{yi2020distributed2} (studying dynamic regret for convex cost and constraint functions).

Algorithm \ref{ag7} depends upon the primal-dual method, but with three differences. 1) $y_{i,t}$ is introduced for each agent $i$ to ultimately track the global information $\sum_{i=1}^N g_{i,t}$ which cannot be directly known to any single agent. 2) $\tilde{\lambda}_{i,t}$ is employed to ensure that $\lambda_{i,t}$'s can eventually achieve consensus for all agents. 3) The term $-\sigma\alpha \tilde{\lambda}_{i,t}$ is added to the update of $\lambda_{i,t}$ for hindering the growth of $\lambda_{i,t}$.

In Algorithm \ref{ag8}, the update of $\tilde{x}_{i,t+1}$ relies on the mirror descent but with an additional regularization function $r_{i,t}$. $b_{i,t}$ is an approximation of $g_{i,t}(\tilde{x}_{i,t+1})$ at the point $x_{i,t}$, and the update of dual variable $\lambda_{i,t}$ is based on gradient ascent with the term $-\beta_t\sum_{j=1}^N w_{ij,t}\lambda_{j,t}$ being used to impede the growth of $\lambda_{i,t}$. As before, $\sum_{j=1}^N w_{ij,t}\lambda_{j,t}$ aims for all agents to achieve an identical dual variable.

Algorithm \ref{ag9} is modified from the primal-dual algorithm, where $\sum_{j=1}^N w_{ij,t}\lambda_{j,t}$ has the same information mixing functionality as used in previous algorithms. The main point is to leverage one-point gradient estimators to replace true gradients in the one-point bandit feedback setting. Note that it is standard to perform the projection on $(1-\vartheta_{i,t})\mathcal{X}_i$, instead of the usual $\mathcal{X}_i$, in this setting, in order to guarantee that $x_{i,t}$ always lies in $\mathcal{X}_i$ after adding $\delta_{i,t}u_{i,t}$. In Algorithm \ref{ag9}, parameters are set as
\begin{align*}
\alpha_{i,t}&=\frac{r_{\mathcal{X}_i}^2}{4mn_i^2F_{g_i}^2t^{\theta_1}},~\beta_{i,t}=\frac{2}{t^{\theta_2}},~\gamma_{i,t}=\frac{1}{t^{1-\theta_2}},\\
\vartheta_{i,t}&=\frac{1}{(t+1)^{\theta_3}},~\delta_{i,t}=\frac{r_{\mathcal{X}_i}}{(t+1)^{\theta_3}},
\end{align*}
where it is assumed $\mathcal{X}_i \supseteq r_{\mathcal{X}_i}\mathcal{B}^{n_i}$ for some $r_{\mathcal{X}_i}>0$, $F_{g_i}>0$ is a constant such that $|[g_{i,t}(x)]_j|\leq F_{g_i}$ for all $x\in\mathcal{X}_i$ with $[\cdot]_j$ being the $j$th component for $j\in[m]$, $\theta_1\in(0,1)$, $\theta_2\in(0,\theta_1/3)$, and $\theta_3\in(\theta_2,(\theta_1-\theta_2)/2]$.

\begin{algorithm}
 \caption{Distributed Primal-Dual Online Learning (DisPDOL) (Full information feedback; Static regret) \cite{li2020onlineLE}}     \label{ag7}
 \begin{algorithmic}[1]
  \STATE {\bf Initialization:} $x_{i,0}\in\mathcal{X}_i$, $\lambda_{i,0}={\bf 0}_m$, $y_{i,0}=Ng_{i,0}(x_{i,0})$.
  \STATE \textbf{Iterations:} Step $t\in[0,T]$, update for each $i\in[N]$:
\begin{align*}
x_{i,t+1}&={\bf P}_{\mathcal{X}_i}[x_{i,t}-\alpha s_{i,t}],      \\
\lambda_{i,t+1}&=\Big[\tilde{\lambda}_{i,t}+\alpha\Big(\sum_{j=1}^N w_{ij,t}y_{j,t}-\sigma\alpha \tilde{\lambda}_{i,t}\Big)\Big]_+,        \\
y_{i,t+1}&=\sum_{j=1}^N w_{ij,t}y_{j,t}+N\big(g_{i,t}(x_{i,t+1})-g_{i,t}(x_{i,t})\big),
\end{align*}
where $s_{i,t}:=\nabla f_{i,t}(x_{i,t})+\nabla g_{i,t}(x_{i,t})^\top\tilde{\lambda}_{i,t}$, $\tilde{\lambda}_{i,t}:=\sum_{j=1}^N w_{ij,t}\lambda_{j,t}$, and the stepsize is set as $\alpha=\sigma^{-1}T^{-\kappa}$ with $\sigma=2N(NB_g^2+1)$, $\kappa\in(0,1)$ and $B_g>0$ being an upper bound on gradients of $f_{i,t}$ and $g_{i,t}$.
 \end{algorithmic}
\end{algorithm}

\begin{algorithm}
 \caption{Distributed Online Primal-Dual Dynamic Mirror Descent (DisOPDM) (Local cost $f_{i,t}+r_{i,t}$ with a regularizer $r_{i,t}$; Static and dynamic regrets) \cite{yi2020distributed}}     \label{ag8}
 \begin{algorithmic}[1]
  \STATE {\bf Input:} Non-increasing sequences $\alpha_t,\beta_t,\gamma_t\in(0,1]$; differentiable and strongly convex functions $\phi_i,i\in[N]$.
  \STATE {\bf Initialization:} $x_{i,0}\in\mathcal{X}_i$ and $\lambda_{i,0}={\bf 0}_m$ for $i\in[N]$.
  \STATE \textbf{Iterations:} Step $t\in[0,T]$, update for each $i\in[N]$:
\begin{align*}
\tilde{x}_{i,t+1}&=\mathop{\arg\min}_{x\in\mathcal{X}_i}\{\alpha_t\langle x,s_{i,t}\rangle+\alpha_t r_{i,t}(x)+\mathcal{D}_{\phi_i}(x,x_{i,t}\},\\
\lambda_{i,t+1}&=\big[\tilde{\lambda}_{i,t}+\gamma_t\big(b_{i,t}-\beta_t\tilde{\lambda}_{i,t}\big)\big]_+,      \\
x_{i,t+1}&=\Phi_{i,t+1}(\tilde{x}_{i,t+1}),
\end{align*}
where $b_{i,t}:=g_{i,t}(x_{i,t})+\nabla g_{i,t}(x_{i,t})(\tilde{x}_{i,t+1}-x_{i,t})$, and $s_{i,t},\tilde{\lambda}_{i,t}$ are the same defined as in Algorithm \ref{ag7}. $\Phi_{i,t}$ is a dynamic model and represents a prior knowledge of the studied problem, which is simply set to be the identity mapping if no any prior knowledge.
 \end{algorithmic}
\end{algorithm}

\begin{algorithm}
 \caption{Distributed Bandit Online Descent with One-Point Sampling Gradient Estimator (DBOD-OSGE) (Dynamic regret) \cite{yi2020distributed2}}     \label{ag9}
 \begin{algorithmic}[1]
  \STATE {\bf Input:} Non-increasing sequences $\alpha_{i,t},\beta_{i,t},\gamma_{i,t}\in(0,1]$; shrinkage parameter $\vartheta_{i,t}\in(0,1)$; $\delta_{i,t}\in(0,r_{\mathcal{X}_i}\vartheta_{i,t}]$ for $i\in[N],t\geq 0$.
  \STATE {\bf Initialization:} $u_{i,0}\in \mathcal{S}^{n_i}$, $z_{i,0}\in (1-\vartheta_{i,0})\mathcal{X}_i$, $x_{i,0}=z_{i,0}+\delta_{i,0}u_{i,0}$, and $\lambda_{i,0}={\bf 0}_m$ for $i\in[N]$.
  \STATE \textbf{Iterations:} Step $t\in[0,T]$, update for each $i\in[N]$:\\
\hspace{0.6cm}Select vector $u_{i,t}\in\mathcal{S}^{n_i}$ independently and uniformly at random.
\begin{align*}
z_{i,t+1}&={\bf P}_{(1-\vartheta_{i,t})\mathcal{X}_i}(z_{i,t}-\alpha_{i,t}s_{i,t}'),    \\
x_{i,t}&=z_{i,t}+\delta_{i,t}u_{i,t},     \\
\lambda_{i,t+1}&=\Big[(1-\beta_{i,t}\gamma_{i,t})\sum_{j=1}^N w_{ij,t}\lambda_{j,t}+\gamma_{i,t}g_{i,t}(x_{i,t})\Big]_+,
\end{align*}
where $s_{i,t}':=\tilde{\nabla}f_{i,t}(z_{i,t})+\tilde{\nabla}g_{i,t}(z_{i,t})^\top\sum_{j=1}^N w_{ij,t}\lambda_{j,t}$, and $\tilde{\nabla}$ means the one-point gradient estimator for functions as defined in (\ref{ope}).
 \end{algorithmic}
\end{algorithm}

\subsection{Online Game}

For online game, there are only a few works in recent years \cite{lu2020online,meng2021decentralized,meng2022decentralized}. Time-invariant coupled inequality constraint is considered in \cite{lu2020online}, which is a special case of time-varying coupled inequality constraints addressed in \cite{meng2021decentralized}. As such, the algorithm proposed in \cite{meng2021decentralized} is provided in Algorithm \ref{ag10}, which is based on primal-dual and mirror descent methods and also extended to the one-point bandit feedback case as well as the case with feedback delays in \cite{meng2022decentralized}. In Algorithm \ref{ag10}, each agent needs to maintain a vector variable $x_{ik,t}\in\mathbb{R}^{n_k}$ to estimate the variable $x_{k,t}$ of agent $k\in[N]$ at time $t$ with $x_{ii,t}:=x_{i,t}$, since in the partial decision information setting, each agent cannot access other agents' decision variables. The term $\sum_{j=1}^N w_{ij}x_{jk,t}$ is designed to achieve consensus of $x_{ik,t}$'s for all agents $i\in[N]$. Moreover, the update of $x_{i,t+1}$ is a weighted average of the previous $x_{i,t}$ and new $\tilde{x}_{i,t+1}$ in order to utilize both last-step and new information. Similar to previous algorithms, $-\beta_t\tilde{\lambda}_{i,t}$ can be employed to hinder the growth of dual variable $\lambda_{i,t}$.

\begin{algorithm}
 \caption{Distributed Online Primal-Dual Dynamic Mirror Descent for OG (DisOPDM-OG) \cite{meng2021decentralized}}     \label{ag10}
 \begin{algorithmic}[1]
  \STATE Each agent $i\in[N]$ maintains a vector $x_{ik,t}\in\mathbb{R}^{n_k}$ for $k\in[N]$ and $\lambda_{i,t}\in\mathbb{R}^m$ at $t$.
  \STATE {\bf Initialization:} $x_{i,0}\in\mathcal{X}_i$, $x_{ik,0}={\bf 0}_{n_k}$ ($k\neq i$), $\lambda_{i,0}={\bf 0}_m$.
  \STATE \textbf{Iterations:} Step $t\in[0,T]$, update for each $i\in[N]$:
{\small\begin{align*}
x_{ik,t+1}&=\sum_{j=1}^N w_{ij}x_{jk,t},~~k\neq i,   \\
\tilde{x}_{i,t+1}&=\mathop{\arg\min}_{x\in\mathcal{X}_i}\big\{\alpha_t\langle x,\nabla_i f_{i,t}({\bf x}_{i,t})+(\nabla g_{i,t}(x_{i,t}))^\top \tilde{\lambda}_{i,t}\rangle   \\
&\hspace{1.7cm}+\mathcal{D}_{\phi_i}(x,x_{i,t})\big\},   \\
x_{i,t+1}&=(1-\alpha_t)x_{i,t}+\alpha_t\tilde{x}_{i,t+1},   \\
\lambda_{i,t+1}&=\Big[\tilde{\lambda}_{i,t}+\gamma_t\Big(g_{i,t}(x_{i,t})-\beta_t\tilde{\lambda}_{i,t}\Big)\Big]_+,
\end{align*}}
where ${\bf x}_{i,t}:=col(x_{i1,t},\ldots,x_{iN,t})$ and $\nabla_i f_{i,t}({\bf x}_{i,t})$ denotes the gradient of $f_{i,t}$ with respect to $x_{i,t}$, $\tilde{\lambda}_{i,t}:=\sum_{j=1}^N w_{ij}\lambda_{j,t}$. And parameters are set as $\alpha_t=1/t^{a_1}$, $\beta_t=1/t^{a_2}$, $\gamma_t=1/t^{1-a_2}$ with $\alpha_0=\beta_0=\gamma_0=1$ and constants $0<a_1<1/2$, $a_1>2a_2$.
 \end{algorithmic}
\end{algorithm}

Finally, it is worth noting that there are many other algorithms for all sorts of different scenarios in the literature, which are too vast to introduce them all. However, all the algorithms for these scenarios can be easily found in the references as discussed in Sections \ref{sc3} and \ref{sc4} below.

\section{Communication Perspective}\label{sc3}

This section is concerned with existing hot topics in DOO and OG from the perspective of information exchanges among agents in the network, which is known to be pivotal to distributed algorithms over multi-agent networks. Without agents' interconnections, it is impossible to tackle the global online learning problem in the distributed manner. In doing so, according to a variety of communication patterns, this section provides a comprehensive review on DOO and OG.


{\em Communication Graphs}. As an important aspect in DOO and OG, the interactions among all the agents in the network are quintessentially captured by graphs. According to different scenarios, it can be mainly classified into four cases, i.e., undirected graphs \cite{yan2012distributed,koppel2015saddle,akbari2018individual,zhang2019distributed,sharma2020distributed}, balanced/unbalanced fixed directed graphs \cite{hosseini2013online,nedic2015decentralized,pang2019randomized,yamashita2021logarithmic,li2021distributedBanUnba,lesage2020dynamic}, balanced time-varying/switching directed graphs \cite{mateos2014distributed,lee2017sublinear,li2018differentially,yuan2020distributed,cesa2020cooperative,yi2020distributed,
yi2020distributed2,li2021distributed2}, and unbalanced time-varying directed graphs \cite{hosseini2016online,lee2017stochastic,akbari2017distributed,zhu2018differentially,lee2016coordinate,li2020distributed}, and random switching graphs \cite{lei2020online,cesa2020cooperative}. Generally speaking, undirected interaction graphs are relatively simple due to the elegant symmetry property of mixing matrix, while time-varying directed graphs, especially unbalanced ones, are more complicated when studying DOO and OG. For example, the authors in \cite{zhang2019distributed} investigated unconstrained DOO for strongly convex objectives, where an improved dynamic regret is established under stationary undirected interconnection graphs. However, it is nontrivial to extend the result in \cite{zhang2019distributed} to the case with unbalanced time-varying directed graphs, since the elegant fixedness and symmetry of the information mixing matrix do not hold under unbalanced time-varying directed graphs.

{\em Full Information Communication}. This is the earliest and most frequently studied case in DOO, where the transmitted information from each agent to its out-neighbors is the agent's full variables, i.e., exact information transmission (e.g., $x_{j,t}$'s in Algorithm \ref{ag1}). Relevant examples include \cite{hosseini2013online,yan2012distributed,mateos2014distributed,koppel2015saddle,zhang2019distributed,han2021differentially,
yuan2017adaptive,sharma2020distributed,yuan2020distributed2,yuan2021distributed,li2020distributed,yi2020distributed,yi2020distributed2,
dixit2020online,yuan2020distributed,shahrampour2017distributed}, and so on, where full information on one or more variables are broadcast to out-neighboring agents. For example, an online distributed dual averaging algorithm was proposed using a sort of regularized projection in \cite{hosseini2013online}, where an exact variable information is propagated in each round. DOO with (strongly) convex functions was addressed in the presence of time-invariant and time-varying coupled inequality constraints in \cite{li2020distributed} and \cite{yi2020distributed}, respectively, where three full variables and one full variable are transmitted, respectively. In addition, a distributed online mirror descent algorithm was developed for DOO with convex functions \cite{shahrampour2017distributed}, where the optimal trajectory consisting of minimizers at all times is assumed to conform with a known linear dynamics corrupted by unknown and unstructured noises.

{\em Privacy-preserving Communication}. Usually, each agent broadcasts its full information to its out-neighboring agents, which, however, is vulnerable to privacy disclosure. Provided that different mass data are distributed over a collection of agents, data privacy for each agent is paramount in a multitude of realistic applications, such as patients' diagnosis information in hospital, personal daily data held on private mobile phone/computer, and users' private information on Facebook, etc. Wherein, private data of each agent are undesirable to be disclosed to other agents. Along this line, an intrinsic privacy-preserving algorithm was proposed for DOO in \cite{yan2012distributed}, where sufficient and necessary conditions were established for privacy preservation, showing that other agents' subgradients (and sensitive raw data) cannot be reconstructed by a malicious learner in networks with greater-than-one connectivity. Furthermore, differential privacy, firstly proposed in \cite{dwork2006differential}, has been taken into consideration in DOO \cite{zhu2018differentially,li2018differentially,xiong2020privacy,lu2020privacy}, where an independent and identically distributed (iid) noise drawn according to Laplace distribution is added to each agent's variable before information transmission. In doing so, it ensures that each agent's exact information cannot be disclosed by other agents due to information masking by Laplace noises. Meanwhile, it is worth noting that except for injected noises, noisy information may be naturally propagated due to the noisy links/channels, or noisy measurements incurred by inexact sensing devices \cite{cao2021decentralized,yang2021distributed}.

{\em Security}. Similar to privacy preservation during communications in multi-agent networks, another critical issue is malicious attacks or adversarial agents, including denial-of-service (DoS) attacks, replay attacks, false data injection attacks, and so on. These attacks can severely vandalize the proposed online algorithms that perform well in benign environments, i.e., in the absence of any intentional attacks. It is well known that designing robust algorithms to adversarial attacks is nontrivial and challenging for multi-agent consensus, distributed learning and distributed optimization, etc. (e.g., \cite{yan2020resilient,yin2018byzantine,liu2021survey}). In this respect, adversarial agents were considered recently for DOO in \cite{sahoo2021distributed}, where Byzantine faulty agents can update its variable arbitrarily, which is then transmitted to its neighbors, with the purpose of preventing no-faulty agents from achieving the optimal solution. And individual static regret is ensured by establishing sufficient conditions on the graph topology, the number and location of the adversarial agents.

{\em Asynchronous Algorithms}. Most of existing works have focused on synchronous algorithms in DOO and OG, that is, all the agents in a network can transmit their information, perform their calculations, and carry out their updating at the same time. As a matter of fact, synchronous algorithms depend on underlying assumptions that a global clock is accessible to all the agents and information transmissions along information channels do not undergo any delayed feedback. Nevertheless, this case may not be true in many practical applications. For example, in wireless sensor networks \cite{sivrikaya2004time}, agents often have different clocks and delayed feedback usually exists during information collection, computation and propagations, thereby incurring distinct time readings at the same global time and different information processing time for each agent, respectively. Along this line, asynchronous algorithms have been extensively investigated for DOO in recent years \cite{cao2021decentralized2,lu2020privacy,jiang2021asynchronous,hsieh2020multi}. Delayed feedback of only local cost function's gradients is addressed in \cite{cao2021decentralized2}. Delayed information transmission among agents is taken into account in \cite{lu2020privacy}. Asymmetric gossiping communication is considered in \cite{jiang2021asynchronous}. And a more general case is studied in \cite{hsieh2020multi}, where inherent delays (the time needed to observe the effect of a decision), computation delays for processing an action at each agent (e.g., gradient computations), and communication delays for information transmission among agents are considered.

{\em Information Quantization or Compression}. One of particular important issues that appear in distributed problems over multi-agent networks is the number of bits transmitted along information links/channels among all the agents. It is because the capacity of transmitted data cannot be arbitrarily large, but subject to physical limitations on transmission channels \cite{gray1998quantization}. Due to this reason, information quantization (or compression), i.e., encoding information with a certain amount of bits, has been extensively studied in multi-agent control (e.g., \cite{li2018quantized,toghani2021scalable}) and centralized/distributed optimization (e.g., \cite{li2020acceleration,magnusson2020maintaining,li2021faster,zhang2021innovation}), etc. However, in contrast to the vast amount of literature in aforementioned fields, to our best knowledge, there are only few existing works focusing on DOO with information quantization \cite{cao2021decentralized,yuan2022distributed}. The proposed algorithms in \cite{cao2021decentralized} leverage only the signs of neighbors' relative variables (a special one of quantization/compression strategies using only one bit), in order to alleviate the sensing and communication requirements for each agent. And random quantization is investigated for distributed online bandit optimization in \cite{yuan2022distributed}, where each agent's random quantized information, instead of full information, is transmitted to its out-neighbors at each time.

\section{Computation Perspective}\label{sc4}

\begin{figure}[H]
\centering
\includegraphics[width=1.8in]{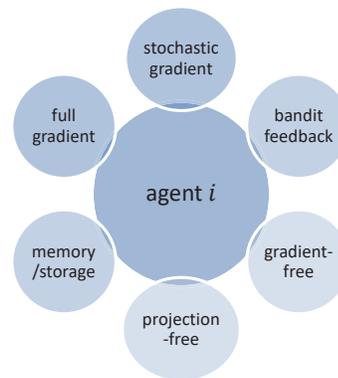}
\caption{Schematic illustration of computation issues for each agent $i\in[N]$ in a network.}
\label{f3}
\end{figure}

In addition to communication aspects discussed in the last section, another important facet for DOO and OG is the computation and memory/storage issues at each agent. As a consequence, this section aims at reviewing specific directions from the perspective of computation, including full gradients/subgradients, stochastic gradients/subgradients, gradient-free methods, projection-free methods, memory/storage requirement, as shown in Fig. \ref{f3}.

{\em Full Gradients/Subgradients}. With regard to this case, full (sub)gradients of local cost functions need to be exactly computed for each agent (e.g., in Algorithm \ref{ag1}), which is relatively easy in some special problems, such as linear cost functions, but usually computationally heavy for general convex and nonconvex functions. Nevertheless, analyzing full gradient case has elegant mathematical theories and can substantially shed light on more complicated scenarios, thus attracting numerous researchers (e.g., \cite{hosseini2013online,yan2012distributed,mateos2014distributed,koppel2015saddle,nedic2015decentralized,
yuan2017adaptive,zhang2019distributed,li2020distributed,yi2020distributed,sharma2020distributed}, to just name a few). Therefore, it has been the most frequently studied case in the literature for DOO and OG until now.

{\em Stochastic Gradients/Subgradients}. It is known that full (sub)gradients are usually computationally prohibitive for general cost and/or constraint functions. Hence, stochastic (sub)gradient methods, i.e., efficiently approximating exact gradients by some data samples, have been increasingly addressed for DOO and OG in recent years \cite{shahrampour2017distributed,lee2017stochastic,zhu2018differentially,li2021distributed2,lu2020privacy}. In the literature, it is usually assumed that stochastic (sub)gradients are unbiased, that is, the mathematical expectation of a stochastic gradient is equal to its true gradient, and have bounded variances.

{\em Gradient-free Methods}. Note that the above stochastic (sub)gradient works often do not provide any concrete approaches on how to calculate stochastic (sub)gradients, instead postulating their unbiasedness and variance boundedness. As two specific instantiations, bandit feedback and sub-optimization solver have been viewed as important gradient-free approaches especially in large-scale learning. To be specific, bandit feedback means that only function values at a few specified points are returned by a calculation oracle. It includes one-point bandit feedback which returns function value at only one point \cite{yuan2021distributed,yi2020distributed2,lei2020online,cao2021decentralized,wang2020push}, and multi-point bandit feedback which reveals function values at multiple points, where two-point bandit feedback is more widely exploited in the literature \cite{yi2020distributed2,yuan2020distributed,pang2019randomized,lei2020online,cao2021decentralized2,li2021distributedBanUnba,li2020onlineLE}. In addition, the sub-optimization solver method connotes that an optimal solution to a sub-optimization problem (often cheaper to resolve than the original problem) can be directly derived, which either has a closed-form solution or can be calculated usually by appealing to sophisticated optimizers, such as (stochastic) gradient descent and so forth. For instance, two sub-optimization problems need to be efficiently solved for the proposed distributed online ADMM algorithm in \cite{akbari2018individual}.

{\em Projection-free Methods}. Another frequently used yet probably computationally heavy operation is to perform projections on feasible closed convex sets, which is only computationally light for several special structured sets, e.g., hyper-box, sphere, and so on. As such, projection-free methods have been considered in DOO with the purpose of alleviating the computational burden for each agent usually by replacing projections with linear optimization over feasible sets \cite{zhang2017projection,wan2020projection}. In this case, the Frank-Wolfe method (or conditional gradient method) is borrowed to develop projection-free distributed online algorithms.

{\em Memory/Storage Requirement}. As local computing is indispensable for each agent, for example, at each updating step, it is imperative for each agent to be capable of storing computing data with a certain amount of data memory. Generally speaking, one common and essential feature in DOO and OG, as in conventional distributed algorithms over multi-agent networks (e.g., distributed optimization), is executing a weighted aggregation or mixing of the received neighboring information, including primal variables, dual variables, and other variables. In order to perform their weighted aggregation operations, one agent requires to store received neighboring information in its buffer, which however can be eliminated once completing this computation. In the meantime, another most common trait is calculating local (sub)gradients, needing to query a (sub)gradient oracle (e.g., full gradient oracle, one- and multi-point bandit oracles) for each agent in each updating round. It should be noted that in some scenarios with gradient-free methods, like the sub-optimization solver methods, gradients' calculation is unnecessary, instead replaced by efficient solutions of sub-optimization problems, such as proximal point algorithm (PPA). In a nutshell, the above mentioned storage and memory are usually the minimal requirements in DOO and OG (e.g., \cite{yan2012distributed,shahrampour2017distributed,yuan2020distributed2,dixit2020online}). On the other hand, besides storing primal variables updated at last round, many works may further require to store other variables, including the widely employed dual variables \cite{koppel2015saddle,akbari2018individual}. Especially for inequality and/or inequality constraints \cite{yuan2017adaptive,sharma2020distributed,yuan2021distributed,yi2020distributed,yi2020distributed2}, it often demands multiple gradient queries, one for local cost function and the others for local/global constrained functions. For example, an additional variable of problem-dimension different from dual variables was introduced and stored for each agent at each round in the literature. It is leveraged to either track the gradients of global cost functions \cite{zhang2019distributed,li2021distributed2}, or assist in mixing neighboring information in order to align them for all agents \cite{mateos2014distributed}, or carry out the averaging of all its own historical decision variables for outputting a desirable decision \cite{hosseini2013online}. Moreover, when addressing unbalanced directed graphs, one more scalar variable is generally necessary to be introduced, stored and transmitted for each agent with the aid of either the push-sum strategy \cite{lee2017stochastic,akbari2017distributed,han2021differentially,li2020distributed} or the balancing weight approach \cite{zhu2018differentially}. Meanwhile, another method for handling the network imbalance is to introduce a network-dimensional variable for each agent to estimate the left eigenvector of the unbalanced interaction matrix \cite{yamashita2021logarithmic}. However, it is memory and computation prohibitive for large-scale networks since the new variable has the same dimension as the network.

\section{State-of-the-art Performance}\label{sec6}

This section is devoted to summarizing various cutting-edge performance results on proposed distributed online algorithms for DOO and OG in the literature, where, to our best knowledge, almost all the existing works have adapted the static and/or dynamic regrets as the performance measures. In doing so, to make various best known results more clear, the overview of state-of-the-art regret bounds is divided into two parts. One is for the case with only simple set constraints and the other is for the case with inequality constraints.

\begin{table*}[ht]
\renewcommand{\arraystretch}{1.3}
\caption{{\upshape State-of-the-art performance results with only simple set constraint $x_i\in\mathcal{X},i\in[N]$ in DOO. ``OB'' stands for optimal bounds, and ``N.F.'' means that the case is not found in the literature.}}
\label{tb1}
\centering
\begin{threeparttable}
\begin{tabular}{c|c|c|c|c}
\hline
Metric & Objective & Full information feedback & One-point bandit feedback & Projection-free algorithms \\
\hline\hline
\multirow{2}*{\makecell[c]{Static\\ regret}} & Convex & \makecell[c]{$O(\sqrt{T})$\\(OB)\\ \cite{yan2012distributed,hosseini2013online,mateos2014distributed,koppel2015saddle,nedic2015decentralized}\\\cite{akbari2018individual,lee2017stochastic}} & \makecell[c]{$O(T^{\frac{3}{4}})$\\ \cite{yuan2021distributed}} & \makecell[c]{$O(T^{\frac{3}{4}})$ \\ with $O(\sqrt{T})$ communication complexity\\ \cite{wan2020projection,wan2021projection}} \\
\cline{2-5}
~ & \makecell[c]{Strongly\\ convex} & \makecell[c]{$O(\log T)$\\(OB)\\ \cite{yan2012distributed,mateos2014distributed,yamashita2021logarithmic}} & \makecell[c]{$T^{\frac{2}{3}}\log T$\\ \cite{yuan2021distributed}} & \makecell[c]{$\tilde{O}(T^{\frac{2}{3}})$\\ with $O(T^{\frac{1}{3}})$ communication complexity\\ (optimal up to logrithmic factors)\\ \cite{wan2021projection}}  \\
\hline
\multirow{2}*{\makecell[c]{(Restricted)\\ Dynamic\\ regret}} & Convex & \makecell[c]{$O(\sqrt{T(1+C_T)})$\\(near OB) \cite{shahrampour2017distributed}\tnote{a}} & N.F. & N.F.  \\
\cline{2-5}
~ & \makecell[c]{Strongly\\ convex} & \makecell[c]{$O(1+P_T)$\\(smooth, near OB) \cite{eshraghi2021improving}\tnote{b}} & N.F. & N.F. \\
\hline
\end{tabular}
  \begin{tablenotes}
        \footnotesize
        \item[a] $C_T=\sum_{t=1}^T\|x_{t+1}^*-A x_t^*\|$ and more details can be found in Algorithm \ref{ag2}.
        \item[b] $P_T:=\sum_{t=2}^T\|x_t^*-x_{t-1}^*\|$ (called {\em path variation/length}, a special case of $V_T^p$ in (\ref{tbn4})) with $x_t^*=\mathop{\arg\min}_{x\in\mathcal{X}}f_t(x)$.
  \end{tablenotes}
\end{threeparttable}
\end{table*}

To this end, the best obtained regret performances for the simple set constraint case are provided from three perspectives: full information feedback, one-point bandit feedback, and projection-free methods, as shown in Table \ref{tb1}. And the best results for the case with inequality constraints are given in Table \ref{tb2}, including uncoupled and coupled inequality constraints, where the uncoupled constraints represent local (or global) inequality constraints gradually revealed to local agents (or all agents) in the network. The employed notations are summarized as follows:
\begin{align}
CACV_T&:=\sum_{t=1}^T\sum_{i=1}^N\sum_{l=1}^m[g^l(x_{i,t})]_+,              \label{tbn1}\\
D\text{-}CCV_T&:=\frac{1}{N}\sum_{i=1}^N\sum_{t=1}^T\|[g_t(x_{i,t})]_+\|,     \label{tbn2}\\
CV_T^+&:=\Big\|\Big[\sum_{t=1}^T g_t(x_t)\Big]_+\Big\|,                        \label{tbn3}\\
V_T^p&:=\sum_{t=2}^T\|\varrho_{t}-\varrho_{t-1}\|,                              \label{tbn4}
\end{align}
where $g^l$ represents the $l$th component of time-invariant constraint functions $g$, i.e., $g=col(g^1,\ldots,g^m)$ with $g^l:\mathbb{R}^n\to\mathbb{R}$, $\{\varrho_t\}_{t=1}^T$ is the sequence of comparators, as introduced when defining the dynamic regret in (\ref{7}), and $V_T^p$ denotes the variation of the sequence of comparators, called {\em path variation/length}.

It should be noted that the case with two-point bandit feedback is not listed in Table \ref{tb1}, since results in this case are generally the same as the full information feedback case. For example, the bounds on static regret, $O(\sqrt{T})$ and $O(\log T)$, were established in \cite{li2021distributedBanUnba} with two-point bandit feedback for convex and strongly convex cost functions, respectively, which are exactly the same as the full information feedback case. They are even the optimal bounds as in centralized online optimization. This fact can be also observed in Table \ref{tb2}, where a few two-point bandit feedback cases are given in \cite{yi2021regret2}. Note that techniques for analyzing the regret performances in Tables \ref{tb1} and \ref{tb2} include some analytical methods in centralized and distributed optimization, some mathematical tricks for dealing with time-varying functions (e.g., the introduction of path variation/length in (\ref{tbn4}), etc.), some technical tricks for constructing special functions to obtain optimal lower bounds on static and dynamic regrets, and so forth.

In what follows, to better discuss the up-to-date performances in Tables \ref{tb1} and \ref{tb2}, let us summarize the state-of-the-art performance results for online optimization in the centralized setting in Table \ref{tb3}, where partial notations can be found in the footnote of Table \ref{tb1} and formulas (\ref{tbn1})-(\ref{tbn4}), and other notations that appear in Table \ref{tb3} are introduced below:
\begin{align}
CCV_T&:=\sum_{t=1}^T \|[g(x_t)]_+\|,                          \label{tbn6}\\
CV_T^l&:=\sum_{t=1}^T g^l(x_t),~~~~~~~~~~\forall l\in[m]         \label{tbn7}\\
V_T^*&:=\sum_{t=1}^T\max_{x\in\mathcal{X}}\|\nabla f_t(x)-\nabla f_{t-1}(x)\|_*^2,         \label{tbn8}\\
V_T^\beta&:=\sum_{t=2}^T t^\beta\|\varrho_t-\varrho_{t-1}\|,~~~\beta\in[0,1)          \label{tbn9}\\
V_T^g&:=\sum_{t=2}^T \sup_{x\in\mathcal{X}}\|g_t(x)-g_{t-1}(x)\|,     \label{tbn10}
\end{align}
\begin{align}
P_T^s&:=\sum_{t=2}^T\|x_t^*-x_{t-1}^*\|^2,                            \label{tbn11}\\
V_T^f&:=\sum_{t=2}^T\sup_{x\in\mathcal{X}}|f_t(x)-f_{t-1}(x)|.         \label{tbn12}
\end{align}

\begin{table*}[!t]
\renewcommand{\arraystretch}{1.3}
\caption{{\upshape State-of-the-art performance results with inequality constraints in DOO. ``IC'' means inequality constraints, ``A-OBF-C'' represents that the result is also Applicable to the case with One-point Bandit Feedback for only Cost functions (but still using gradients of constraint functions), ``A-TBF-C2'' signifies that the result is also Applicable to the case with Two-point Bandit Feedback for both Cost and Constraint functions, ``OBF-C'' (resp. ``OBF-C2'') means that the result is derived under One-point Bandit Feedback for only Cost functions (resp. both Cost and Constraint functions), and ``SC'' denotes the result derived under Slater's condition.}}
\label{tb2}
\centering
\begin{threeparttable}
\begin{tabular}{c|c|c|c|c|c}
\hline
\multirow{2}*{Metric} & \multirow{2}*{Objective} & \multicolumn{2}{c|}{Local inequality constraints} & \multicolumn{2}{c}{Coupled inequality constraints}  \\
\cline{3-6}
~ & ~ & Time-invariant & Time-varying & Time-invariant & Time-varying \\
\hline\hline
\multirow{2}*{\makecell[c]{Static\\ regret}} & Convex & \makecell[c]{$O(T^{\frac{3}{4}})$\\$CACV_T=O(T^{\frac{5}{8}})$\\ (A-OBF-C) \cite{yuan2021distributed}\tnote{1}} & \makecell[c]{$O(\sqrt{T})$ \\ $D\text{-}CCV_T=O(T^{\frac{3}{4}})$\\(A-TBF-C2) \cite{yi2021regret2}\tnote{1}} & N.B.\tnote{2} & \makecell[c]{$O(\sqrt{T})$\\ $CV_T^+=O(T^{\frac{3}{4}})$\\ \cite{li2020onlineLE}} \\
\cline{2-6}
~ & \makecell[c]{Strongly\\ convex} & \makecell[c]{$O(\log T)$\\$CACV_T=O(\sqrt{T\log T})$ \cite{yuan2021distributed}\\and\\$O(T^{\frac{2}{3}}\log T)$ \\$CACV_T=O(\sqrt{T\log T})$\\ (OBF-C) \cite{yuan2021distributed}} & \makecell[c]{$O(T^\kappa)$\\ $D\text{-}CCV_T=O(T^{1-\frac{\kappa}{2}})$\\ $\kappa\in(0,1)$\\ (A-TBF-C2) \cite{yi2021regret2}} & N.B. & \makecell[c]{$O(T^\kappa)$\\$CV_T^+=O(T^{1-\frac{\kappa}{2}})$ or\\$CV_T^+=O(T^{\max\{\kappa,1-\kappa\}})$ (SC)\\$\kappa\in(0,1)$ \cite{yi2020distributed}} \\
\hline
\multirow{2}*{\makecell[c]{(Restricted)\\ Dynamic\\ regret}} & Convex & N.F.\tnote{3} & \makecell[c]{$O(T^{1-\kappa}+T^\kappa(1+V_T^p))$ \\ $D\text{-}CCV_T=O(T^{1-\frac{\kappa}{2}})$\\ $\kappa\in(0,1)$\\ (A-TBF-C2) \cite{yi2021regret2}} & N.B. & \makecell[c]{$O(\max\{T^{\max\{\kappa,1-\kappa\}},T^\kappa P_T\})$\\$CV_T^+=O(T^{1-\frac{\kappa}{2}})$ or\\ $CV_T^+=O(T^{\max\{\kappa,1-\kappa\}}$ (SC)\\ $\kappa\in(0,1)$ \cite{yi2020distributed}\\and \\$O(\max\{T^\theta,T^\theta V_T^p\})$\\$\mathbb{E}(CV_T^+)=O(T^{\frac{7}{4}-\theta})$\\$\theta\in(\frac{3}{4},1)$ (OBF-C2)\cite{yi2020distributed2}} \\
\cline{2-6}
~ & \makecell[c]{Strongly\\ convex} & N.F. & N.F. & N.F. & N.F. \\
\hline
\end{tabular}
  \begin{tablenotes}
        \footnotesize
        \item[1] Note that $CACV_T$ and $D\text{-}CCV_T$ are strictly tighter metrics than $CV_T^+$ defined in (\ref{tbn3}) in general.
        \item[2] ``N.B.'' means no better results than the corresponding time-varying coupled inequality constraint case (note that time-invariant constraints are a special case of time-varying constraints).
        \item[3] ``N.F.'' means that the case is not found in the literature.
  \end{tablenotes}
\end{threeparttable}
\end{table*}

From Table \ref{tb1}, it is easy to see that the static regret bounds in full information feedback are $O(\sqrt{T})$ and $O(\log T)$ for convex and strongly convex cost functions, respectively, which are the same as the optimal/best ones in centralized online optimization as shown in Table \ref{tb3}. That is, the results in full information feedback are already the best ones. When gradients are not available in one-point bandit feedback, where the one-point gradient estimator is employed by relying on only a one-point function value, it can be found that the performance on static regrets is worse than the full information case for both convex and strongly convex cost functions. However, the static regret bounds in this case are still worse than the near-optimal result $\tilde{O}(\min\{\sqrt{nT},T^{\frac{3}{4}}\})$ \cite{saha2021optimal} in the centralized setting. In this regard, it is still open how to further improve the static regret bounds in the distributed setup. Similarly, when reducing the computational complexity by leveraging projection-free algorithms (e.g., Frank-Wolf) without projection operations, the static regret bounds are also inferior to the full information feedback case since the projection-free algorithm cannot accurately compute the gradient in general. However, it is noteworthy that the current results are almost the same as the best obtained results in the centralized scenario (cf. Table \ref{tb3}), although it is still open if they are the optimal ones. Moreover, the one-point bandit feedback and projection-free cases generally perform worse than the full information case, but it is hard to distinguish which one is better between the one-point bandit feedback and projection-free scenarios. Finally, from Table \ref{tb1}, the dynamic regret results are near-optimal for both convex and strongly convex cost functions, while other two scenarios are still lacking in the literature.

For scenarios with inequality constraints in the distributed setting, most of results in Table \ref{tb2} for DOO are generally inferior to the corresponding best-known results in centralized setting as summarized in Table \ref{tb3}, mainly due to the complexity of local information communications among all agents. From Table \ref{tb2}, it is easy to observe that static regret bounds for strongly convex cost functions are better than corresponding scenarios for convex cost functions, which is consistent with the intuition. Generally speaking, time-varying constraints are more challenging than fixed constraints under both local and coupled constraint settings, and coupled constraints are more complicated than local constraints due to the unavailability of some information on coupled constraints. Note that local constraints mean that constraints are solely for each individual agent, i.e., being irrelevant to all other agents. Also, the results for time-varying constraints can be also applied to handle fixed constraints which can be regarded as a special case of time-varying ones, but not vice versa. Note that the performance metrics $CACV_T$ and $D\text{-}CCV_T$ for inequality constraints are more stringent than $CV_T^+$, since the projection operation $[\cdot]_+$ is directly performed on constraint functions, instead of the sum of them over the time horizon $t\in[T]$. In summary, the scenarios with inequality constraints have been less investigated than the simple set constraint case, and further study is needed to improve the performances. Even in the centralized setting with inequality constraints, as shown in Table \ref{tb3}, it is still unclear for many best-known results if they are optimal ones.

{\em Accelerated and Adaptive Methods.} Even though several regret bounds are optimal in terms of $T$, as in centralized/distributed optimization, there are also some methods to improve the regret performance or reduce the conditions while maintaining the same performance. For instance, the near-optimal bound $O(1+P_T)$ on dynamic regret is usually obtained for strongly convex functions. However, it was derived for convex functions in \cite{lesage2020dynamic}, but under a stronger assumption on the communication graph, i.e., every agent is connected to at least two other agents in the network. Moreover, some accelerated and improved methods, such as Nesterov accelerated gradient method and adaptive gradient methods, have been leveraged to further improve the performance. For example, the momentum acceleration technique was exploited to improve the performance under time-varying unbalanced communication graphs in \cite{fang2021accelerated}, where an improved static regret bound $O(\sqrt{1+\log T}+\sqrt{T})$ is established for convex cost functions. Also, adaptive gradient methods have been integrated into DOO in \cite{nazari2019adaptive,nazari2019dadam,carnevale2020distributed}.

\begin{table*}[!t]
\renewcommand{\arraystretch}{1.3}
\caption{{\upshape State-of-the-art performance results in centralized online optimization. To save the space, the following abbreviations are made: ``IC'' is the abbreviation of inequality constraints, ``BCO'' means bandit convex optimization, ``SC'' denotes the result derived under Slater's condition, ``OB'' stands for optimal bounds, ``EBC'' means the result obtained under error bound condition (weaker than SC \cite{wei2020online}), and ``N.F.'' means that the case is not found in the literature.}}
\label{tb3}
\centering
\begin{threeparttable}
\begin{tabular}{c|c|c|c|c|c|c}
\hline
Metric & Objective & Full information feedback & One-point bandit & Projection-free & Fixed IC & Time-varying IC\\
\hline\hline
\multirow{2}*{\makecell[c]{Static\\ regret}} & Convex & \makecell[c]{$O(\sqrt{T})$\\(OB) \cite{cesa2006prediction}} & \multirow{2}*{\makecell[c]{$\tilde{O}(\min\{\sqrt{nT},T^{\frac{3}{4}}\})$ \\ (pseudo-1d BCO) \\(OB, up to \\logarithmic factors)\\ \cite{saha2021optimal}\tnote{a}}} & \makecell[c]{$O(T^{\frac{3}{4}})$\\(nonsmooth)\\\cite{hazan2012projection}; \\ $O(T^{\frac{2}{3}})$\\(smooth)\\ \cite{hazan2020faster}} & \makecell[c]{$O(\sqrt{T})$ \\ $CCV_T=O(T^{\frac{1}{4}})$ \cite{yi2021regret};\\or\\$O(\sqrt{T})$, $CV_T^l=O(1)$ \\ (SC, OB) \cite{yu2020low,liu2020pond};\\or\\$O(\max\{V_T^*,L_f\})$\tnote{b}\\(SC) \cite{qiu2020beyond}} & \makecell[c]{$O(\sqrt{T})$\\$CV_T^l=O(\sqrt{T})$\\(SC) \cite{neely2017online}\\or\\$O(\sqrt{T})$\\$\mathbb{E}(CV_T^+)=O(\sqrt{T})$\\(i.i.d. constraints,\\unknown distribution)\\(EBC) \cite{wei2020online}} \\
\cline{2-3}\cline{5-7}
~ & \makecell[c]{Strongly\\ convex} & \makecell[c]{$O(\log T)$\\(OB) \cite{hazan2014beyond}} &   & \makecell[c]{$O(T^{\frac{2}{3}})$\\ \cite{wan2021projection,garber2021revisiting}} & \makecell[c]{$O(\log T)$\\$CCV_T=O(\log T)$\\ \cite{yi2021regret}} & N.F. \\
\hline
\multirow{2}*{\makecell[c]{Dynamic\\ regret}} & Convex & \makecell[c]{$O(P_T)$\\(smooth) \cite{yang2016tracking}\tnote{c};\\ $O(\sqrt{T}+\sqrt{T^{1-\beta}V_T^\beta})$\\ (OB) \cite{zhao2021proximal}} & \makecell[c]{$O(T^{\frac{3}{4}}+T^{\frac{1}{4}}P_T)$\\ \cite{li2021online}} & N.F. & \makecell[c]{$O(\sqrt{T}(1+V_T^p))$\\$CCV_T=O(T^{\frac{1}{4}})$ \cite{yi2021regret};\\or\\
$O(\sqrt{T(1+V_T^p)})$\\$CCV_T=O(\sqrt{T})$ \cite{yi2021regret}} & \makecell[c]{$O(\max\{\sqrt{T},\sqrt{T}P_T\})$\\$CV_T^l=O(\sqrt{T})$ (SC) or\\$CV_T^l=O(T^{\frac{3}{4}})$ \cite{ding2021dynamic}\\or\\$O(\max\{\sqrt{TP_T},V_T^g\})$\\$CV_T^l=$\\$O(\max\{\sqrt{T},V_T^g\})$ \cite{fu2021elastic}\\or\\$O(\sqrt{TP_T})$\\$CV_T^l=$\\$O(\max\{T^{\frac{3}{4}},V_T^g\})$ \cite{fu2021elastic}}  \\
\cline{2-7}
~ & \makecell[c]{Strongly\\ convex} & \makecell[c]{$O(\min\{P_T,P_T^s,V_T^f\})$\\ (smooth, OB) \cite{zhao2021improved}} & N.F. & N.F. & N.F. & N.F.\\
\hline
\end{tabular}
  \begin{tablenotes}
        \footnotesize
        \item[a] This result improves a lower bound of $O(n\sqrt{T})$ even for strongly convex and smooth cost functions in \cite{shamir2013complexity}.
        \item[b] $L_f$ is the Lipschitz constant for $\nabla f_t$.
        \item[c] An additional assumption is imposed in \cite{yang2016tracking}, that is, the optimal decision $x_t^*\in \mathcal{X}$ at each time $t$ is global (i.e., $\nabla f_t(x_t^*)={\bf 0}$).
  \end{tablenotes}
\end{threeparttable}
\end{table*}

Note that all the above discussed works are in the setting of cooperative agents. For noncooperative agents in online game, the existing works are summarized in Table \ref{tb4}. In this case, {\em individual/local dynamic regret} for online game is defined as
\begin{align}
D\text{-}Reg_i^g:=\sum_{t=1}^T f_{i,t}(x_{i,t},x_{-i,t}^*)-\sum_{t=1}^T f_{i,t}(x_{i,t}^*,x_{-i,t}^*),        \label{og1}
\end{align}
where $x_t^*=col(x_{i,t}^*,x_{-i,t}^*)$ is a (generalized) Nash equilibrium of online game at round $t$. Similarly, {\em individual/local static regret} for online game can be defined as
\begin{align}
S\text{-}Reg_i^g:=\sum_{t=1}^T f_{i,t}(x_{i,t},x_{-i}^*)-\sum_{t=1}^T f_{i,t}(x_{i}^*,x_{-i}^*),        \label{og2}
\end{align}
where $x^*=col(x_i^*,x_{-i}^*)$ is a (generalized) Nash equilibrium of an offline game with $\sum_{t=1}^T f_{i,t}(x_i,x_{-i})$ being the cost function of agent $i$ for all $i\in[N]$.

It can be found from Table \ref{tb4} that the results under fixed inequality constraints \cite{lu2020online} are better than those under time-varying inequality constraints \cite{meng2021decentralized,meng2022decentralized}, since time-varying scenarios are more challenging. Moreover, the results in one-point bandit feedback are worse than those in full information feedback, which conforms with our intuition as discussed before. Overall, it is unknown what the optimal results are for these scenarios, especially for the performance metric $CV_T^+$ on inequality constraints, and more efforts are required to derive better or optimal results in future.

It is worth noticing that the benchmark used in (\ref{og1}) and (\ref{og2}) is Nash equilibria. Other alternatives can be defined as
\begin{align}
D\text{-}R_i^{g}&:=\sum_{t=1}^T \big[f_{i,t}(x_{i,t},x_{-i,t})-\min_{y_{i,t}\in\mathcal{X}_{i,t}}f_{i,t}(y_{i,t},x_{-i,t})\big],    \nonumber\\
S\text{-}R_i^{g}&:=\sum_{t=1}^T f_{i,t}(x_{i,t},x_{-i,t})-\min_{y_i\in\mathcal{X}_{i,t}}\sum_{t=1}^T f_{i,t}(y_{i},x_{-i,t}),       \nonumber
\end{align}
among which the second one has been extensively studied for no-regret learning of stationary games, i.e., games with time-invariant cost functions. It is worth noting that the sublinearity of regrets $D\text{-}R_i^g$ and $S\text{-}R_i^g$, in general, cannot ensure the convergence to a Nash equilibrium even for stationary games \cite{mertikopoulos2018cycles,vlatakis2020no}.

\begin{table*}[!t]
\renewcommand{\arraystretch}{1.3}
\caption{{\upshape State-of-the-art performance results in online game.}}
\label{tb4}
\centering
\begin{threeparttable}
\begin{tabular}{c|c|c|c|c}
\hline
\multirow{3}*{Metric} & \multirow{3}*{Objective} & \multicolumn{3}{c}{Coupled inequality constraints}  \\
\cline{3-5}
~ & ~ & \multirow{2}*{\makecell[c]{Time-invariant\\(Full information feedback)}} & \multicolumn{2}{c}{Time-varying} \\
\cline{4-5}
~ & ~ & ~ & Full information feedback & One-point bandit feedback \\
\hline\hline
\makecell[c]{Individual\\dynamic regret} & Strong monotonicity & \makecell[c]{$O(\sqrt{T(1+P_T)})$\\ $CV_T^+=O(\sqrt{T(1+P_T)})$\\ \cite{lu2020online}\tnote{c}} & \makecell[c]{$O(T^{\frac{5}{6}}+T^{\frac{5}{6}}\sqrt{P_T})$\\ $CV_T^+=O(T^{\frac{5}{6}}+T^{\frac{5}{6}}\sqrt{P_T})$\\ \cite{meng2021decentralized}} & \makecell[c]{$O(T^{\frac{13}{14}}+T^{\frac{13}{14}}\sqrt{P_T})$\\ $\mathbb{E}(CV_T^+)=O(T^{\frac{13}{14}}+T^{\frac{13}{14}}\sqrt{P_T})$\\ \cite{meng2022decentralized}} \\
\hline
\end{tabular}
  \begin{tablenotes}
        \footnotesize
        \item[c] $P_T$ is defined in the footnote of Table \ref{tb1}. However, $x_t^*,t\in[T]$ used in $P_T$ is slightly different here, denoting the Nash equilibrium (instead of a minimizer) at time instant $t$.
  \end{tablenotes}
\end{threeparttable}
\end{table*}

\section{Numerical Experiments}

\begin{figure}[H]
\centering
\includegraphics[width=2.5in]{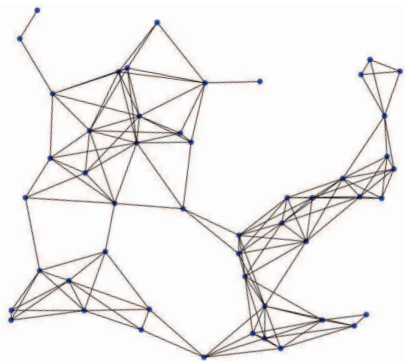}
\caption{The communication graph randomly generated among $N=50$ agents.}
\label{sim-graph}
\end{figure}

\begin{figure}[H]
\centering
\includegraphics[width=2.9in]{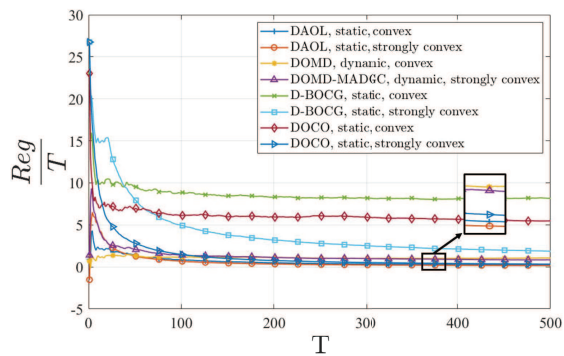}
\caption{Numerical results on Algorithms \ref{ag1}-\ref{ag4} and \ref{ag5}, where DOCO means Algorithm \ref{ag5} without inequality constraints, and $Reg$ denotes a generic regret (either static or dynamic regret) depending on different metrics considered in different algorithms. ``Static'' and ``dynamic'' represent which regret is leveraged.}
\label{sim1}
\end{figure}

This section provides some numerical examples to show and compare the efficacy of Algorithms \ref{ag1}-\ref{ag10}. To do so, let us consider the distributed online (regularized) linear regression problem over a network consisting of $N=50$ agents \cite{yuan2020distributed,yuan2021distributed}, formulated as
\begin{align}
f_t(x)=\sum_{i=1}^N f_{i,t}(x)=\sum_{i=1}^N \frac{1}{2}\big(o_{i,t}^\top x-l_{i,t}\big)^2,~~x\in\mathcal{X}      \label{ne1}
\end{align}
where the data $(o_{i,t},l_{i,t})\in\mathbb{R}^n\times\mathbb{R}$ is revealed only to agent $i$ at time $t$. Note that the function in (\ref{ne1}) is convex, which will be added a regularizer $\theta\|x\|^2$ when considering strongly convex functions for some algorithms, where $\theta>0$ is a regularization parameter. In the simulations, let $\mathcal{X}=\{x\in\mathbb{R}^n|\|x\|\leq 1\}$, and set $\theta=0.1$ and $n=3$. To better compare different algorithms, the same randomly generated undirected communication graph in Fig. \ref{sim-graph} is employed for all algorithms. And functions $\phi,\phi_i$ in Bregman divergence are all selected as $\|x\|^2/2$ for all relevant algorithms. Moreover, each entry of $o_{i,t}$ is generated from the interval $(-1,1)$ uniformly, and the response is given by
\begin{align}
l_{i,t}=\langle o_{i,t},b\rangle +\epsilon_{i,t},     \label{ne2}
\end{align}
where the $i$th entry of $b\in\mathbb{R}^n$ equals $1$ for $1\leq i \leq \lfloor \frac{n}{2}\rfloor$ and 0 otherwise. In addition, the noise $\epsilon_{i,t}$ is generated from the normal distribution $\mathcal{N}(0,1)$ in an independent and identically distributed manner. Then, with the same or similar parameters used in these algorithm design, the numerical results on Algorithms \ref{ag1}-\ref{ag4} and \ref{ag5} without inequality constraints (i.e., DOCO) are shown in Fig. \ref{sim1}.

It can be observed from Fig. \ref{sim1} that all algorithms are efficient, and it is not surprising to see that the performance for strongly convex cost functions is better than that of the same algorithm for convex ones. Furthermore, it is shown that Algorithm \ref{ag1} with full information based on gradient descent outperforms other algorithms, especially projection-free Algorithm \ref{ag4} and Algorithm \ref{ag5} with one-point bandit feedback. This is consistent with the pervasive intuition. In addition, it is found that Algorithm \ref{ag5} performs better than Algorithm \ref{ag4}, although they have similar theoretical results, indicating that it is promising to derive better performances for the case with one-point bandit feedback.

\begin{figure}[H]
\centering
\includegraphics[width=2.9in]{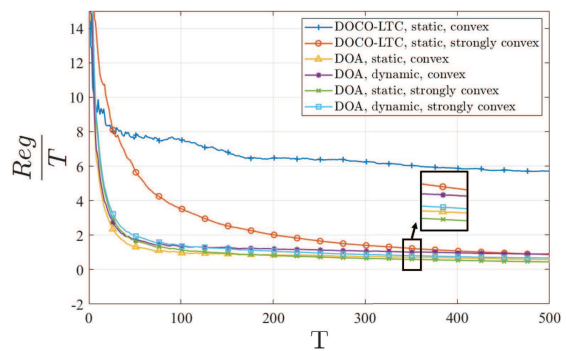}
\caption{Numerical results on regrets for Algorithms \ref{ag5} and \ref{ag6}.}
\label{sim2-1}
\end{figure}

\begin{figure}[H]
\centering
\includegraphics[width=2.9in]{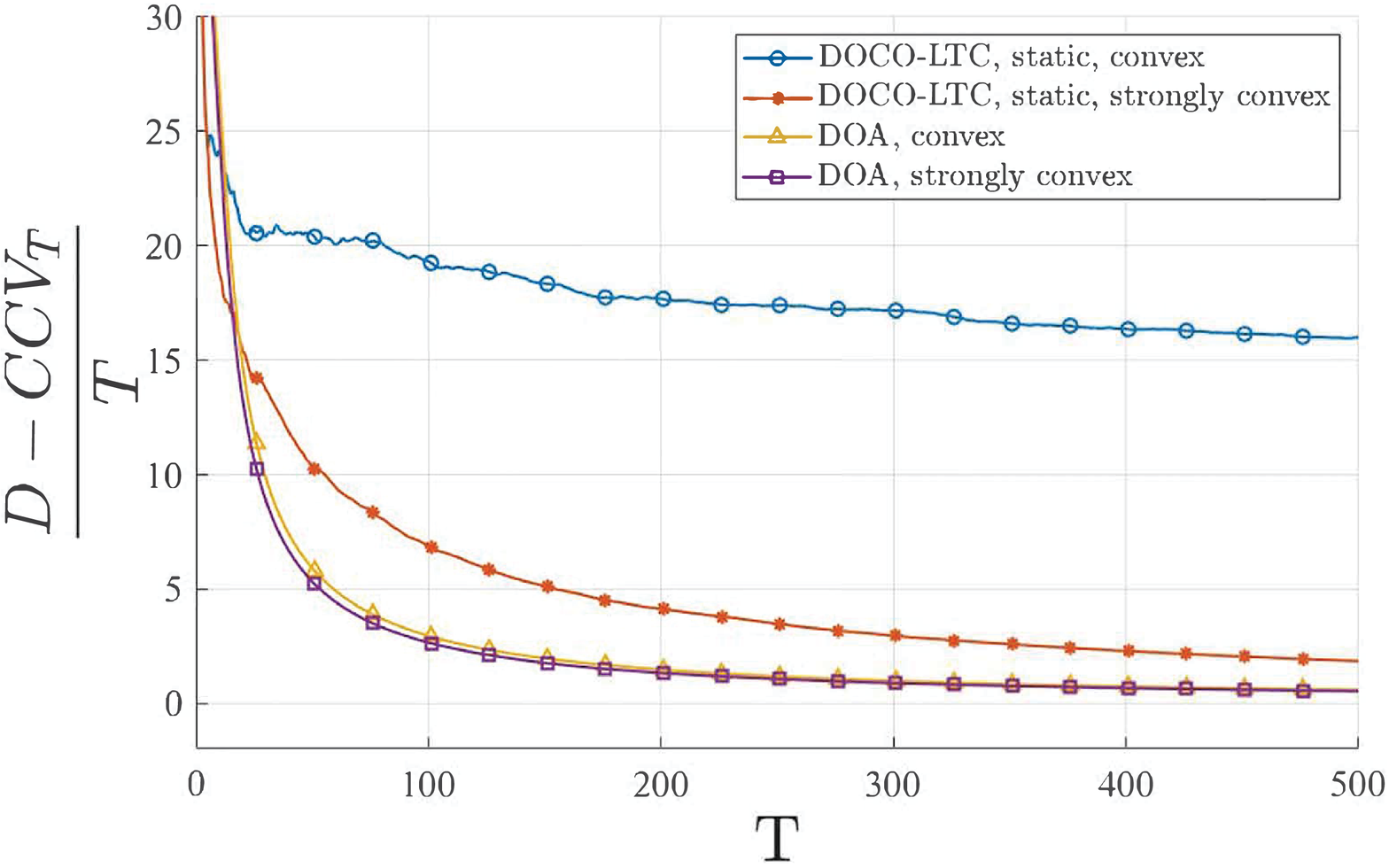}
\caption{Numerical results on inequality constraint $D\text{-}CCV_T$ for Algorithms \ref{ag5} and \ref{ag6}.}
\label{sim2-2}
\end{figure}

For Algorithms \ref{ag5} and \ref{ag6} with local inequality constraints, similar to \cite{yuan2021distributed}, the local constraint functions are all given as $g_{i,t}(x)=x-a$ for some $a\in\mathbb{R}^n$. By setting $a=col(0.1,0.1,0.1)$, running Algorithms \ref{ag5} and \ref{ag6} gives rise to the results on regrets and inequality constraints in Figs. \ref{sim2-1} and \ref{sim2-2}, respectively. It can be observed that the performance with full information feedback (Algorithm \ref{ag6}) is better than that with one-point bandit feedback (Algorithm \ref{ag5}), thus supporting our intuition and theoretical results.

\begin{figure}[H]
\centering
\includegraphics[width=2.9in]{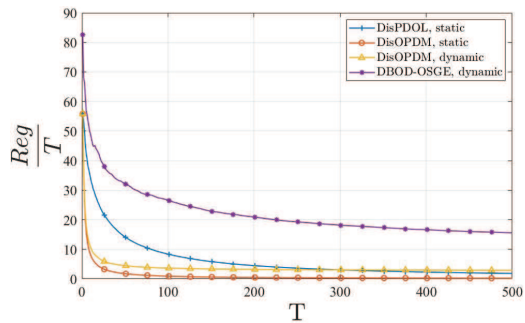}
\caption{Numerical results on regrets for Algorithms \ref{ag7}-\ref{ag9}.}
\label{sim3-1}
\end{figure}

\begin{figure}[H]
\centering
\includegraphics[width=2.9in]{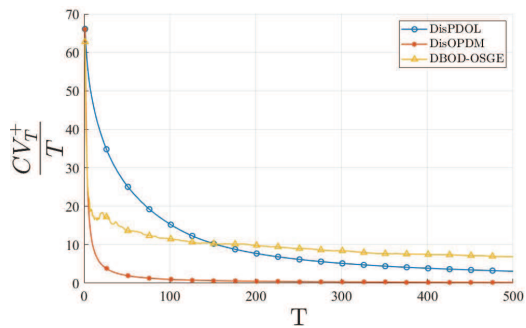}
\caption{Numerical results on inequality constraint $CV_T^+$ for Algorithms \ref{ag7}-\ref{ag9}.}
\label{sim3-2}
\end{figure}

For Algorithms \ref{ag7}-\ref{ag9} with coupled inequality constraints, consider the plug-in electric vehicle (PEV) charging problem as in \cite{li2020distributed,li2020onlineLE} with local cost and constraint functions being of the form
\begin{align}
f_{i,t}(x_i)&=\frac{\bar{a}_{i,t}}{2}\|x_i\|^2+\bar{b}_{i,t}^\top x_i,     \label{ne3-1}\\
g_{i,t}(x_i)&=x_i+d_{i,t},~~~~x_i\in\mathcal{X}_i\subseteq\mathbb{R}^{n_i}   \label{ne3-2}
\end{align}
where $\bar{a}_{i,t}>0$ and $\bar{b}_{i,t}\in\mathbb{R}^{n_i}$ are time-varying, indicating dynamic charging cost for charging PEV at different times. Let $n_i=1$ for all $i\in[N]$, $\Phi_i$ be the identity mapping in Algorithm \ref{ag7}, $r_{i,t}\equiv 0$ in Algorithm \ref{ag8}, and $\bar{a}_{i,t},\bar{b}_{i,t},d_{i,t}$ be generated uniformly from intervals $(0.5,1.5)$, $(0,1)$, $(0.2,0.5)$, respectively. The numerical results are given in Figs. \ref{sim3-1} and \ref{sim3-2}.

It is easy to see that Algorithm \ref{ag9} has the worst performance due to the least information it has in the one-point bandit feedback scenario. From Fig. \ref{sim3-1}, one can observe that the static regret performance is better than the dynamic one for DisOPDM, since the dynamic regret is generally greater than the static one as seen from their definitions. Finally, DisOPDM is shown to be superior to DisPDOL, indicating that $b_{i,t}$ in Algorithm \ref{ag8} outperforms $\sum_{j=1}^N w_{ij,t}y_{j,t}$ in Algorithm \ref{ag7} employed to update dual variables.

\begin{figure}[H]
\centering
\includegraphics[width=2.6in]{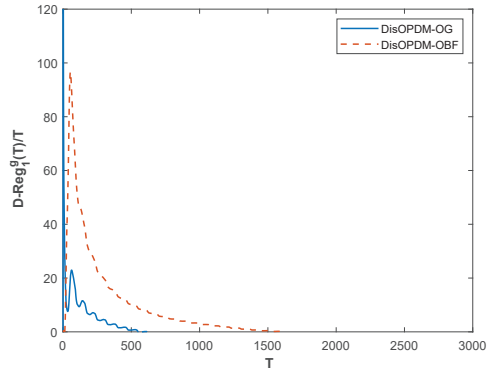}
\caption{Numerical results on regret for Algorithm \ref{ag10} and its one-point bandit variant DisOPDM-OBF in \cite{meng2022decentralized}.}
\label{sim4-1}
\end{figure}

\begin{figure}[H]
\centering
\includegraphics[width=2.6in]{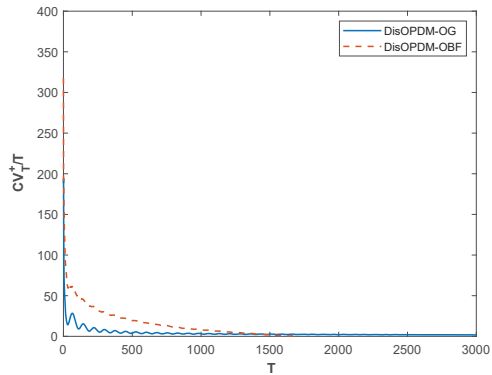}
\caption{Numerical results on inequality constraint $CV_T^+$ for Algorithm \ref{ag10} and its one-point bandit variant DisOPDM-OBF in \cite{meng2022decentralized}.}
\label{sim4-2}
\end{figure}

For online game, Algorithm \ref{ag10} and its one-point bandit variant (called DisOPDM-OBF) in \cite{meng2022decentralized} are tested by online Nash-Cournot game as given by Example \ref{em3} in Subsection \ref{s2-ss4}. As in \cite{meng2022decentralized}, let us set $x_{i,t}\in\mathcal{X}_i=[0,30]$, the production cost $p_{i,t}(x_{i,t})=x_{i,t}(sin(t/12)+1)$, and the firm $i$'s demand price $d_{i,t}(x_t)=21+i/9-0.5i*\sin(t/12)-\sum_{k=1}^N x_{k,t}$. Moreover, as the time-varying market capacity constraint, the coupled inequality constraint is $\sum_{i=1}^N x_{i,t}\leq\sum_{i=1}^N l_{i,t}$ with $l_{i,t}=10+\sin(t/12)$ being the local bound only accessible to firm $i$. Then, Figs. \ref{sim4-1} and \ref{sim4-2} present the simulation results, where only result on the first firm is displayed in order to clearly observe the difference between the algorithms, but other firms have similar results. Both figures show that the full information feedback is better than the one-point bandit feedback, being in accordance with the intuition as mentioned earlier.

\section{Future Directions}

With the above discussions on DOO and OG, it can be found that the research on DOO and OG is still in its infancy. To facilitate further studies in this domain for both beginners and specialists, this section aims at pointing out possible future research directions in DOO and OG.

{\em Better Performance.} By comparing existing performance results on DOO in Tables \ref{tb1} and \ref{tb2} with the centralized best-known or optimal results in Table \ref{tb3}, a multitude of gaps still need to be bridged for both DOO and OG. For example, the static regret bound in the distributed case with one-point bandit feedback is worse than the optimal bound as obtained in the centralized setup, and in the case with uncoupled/coupled inequality constraints, lots of regret bounds are also inferior to the best known results in the centralized setting. However, it is unclear if all the gaps can be exactly bridged, although existing results have shown that some gaps can be indeed bridged in some scenarios (e.g., static regret bounds in full information feedback). It is still open but promising for researchers to bridge these gaps. Moreover, there are a number of open issues in DOO and OG to be addressed. As an example, the dynamic regret bound for DOO with one-point bandit feedback is still vacant.

{\em Network Effect.} One of important issues in distributed algorithms is to consider the effect of network, such as the agent number $N$ and other network parameters, on the algorithms' performances. In this respect, a dynamic regret bound for convex cost functions is established in \cite{shahrampour2017distributed} as
      \begin{align}
      O\bigg(\sqrt{\frac{NT(1+C_T)}{1-\lambda_2(W)}}\bigg),       \label{fd1}
      \end{align}
      where the parameters can be found in Algorithm \ref{ag2}. It is easy to see that the regret bound is proportional to $\sqrt{N}$ and inversely proportional to $\sqrt{1-\lambda_2(W)}$. In addition, a dynamic regret bound, i.e.,
      \begin{align}
      GP_T+O\bigg(\frac{NG^2}{\gamma(1-\gamma^{1/\zeta})}+\frac{N^2G^2}{1-w_{max}}\bigg),       \label{fd2}
      \end{align}
is derived in \cite{lesage2020dynamic} for convex functions under a stronger assumption on the communication graph. That is, every agent is connected to at least two other agents in the network \cite{lesage2020dynamic}, where $\gamma\in[0,1),\zeta\geq 1$ are graph-related parameters, $w_{max}:=\max_{i,j\in{N}}w_{ij}$, and $G$ is a constant such that $\|\nabla f_{i,t}(\cdot)\|_*\leq G$. One can observe that the bound is proportional to $N^2$ and inversely proportional to some graph parameters. The above bounds are neat results in the literature. However, it is still unclear whether the dependence on those graph parameters are sharp or not.

{\em Information Quantization/Compression.} It is already known that the transmission capacity along information channels among neighboring agents is paramount for distributed algorithms over multi-agent networks. Therefore, information quantization/compression for both transmitted message and local computations (e.g., gradient calculation) is imperative in future research directions. The current relevant research is still lacking in DOO and OG (two exceptions are \cite{cao2021decentralized,yuan2022distributed}), although it has been extensively considered in multi-agent control (e.g., \cite{li2018quantized,toghani2021scalable}) and centralized/distributed optimization (e.g., \cite{li2020acceleration,magnusson2020maintaining,li2021faster,zhang2021innovation}), etc.

{\em DOO and OG with Control Systems.} Most of existing works have focused on distributed online learning in the absence of system dynamics, that is, without considering physical control systems, which can be viewed as information-layer problems. Nevertheless, an agent often has its physical operating dynamics (e.g., bicycle dynamics for robots), which should be appropriately considered and controlled, thought of as physical layer problems. Although recent research \cite{chang2021regret,yu2022continuous} has integrated the control system dynamics into DOO, the related research is yet to be fully explored in order to smoothly apply distributed online algorithms to real-world problems.

{\em Continuous-time Algorithms.} Most of proposed algorithms in DOO and OG are discrete-time iterated mainly due to the discrete-time computation fact of realistic implementations, for example, by computers. Even so, many physical systems or phenomena in practice are in continuous-time domain, such as the continuous-time operating dynamics of electric current flow and so forth. On the other hand, continuous-time algorithms can also be a powerful tool for providing insight into discrete-time algorithms (e.g., \cite{muehlebach2021optimization,sanz2021connections}). Along this line, the continuous-time setup has been addressed with local inequality constraints \cite{paternain2020distributed} and coupled inequality constraints \cite{yu2022continuous}. However, this type of problems are still less addressed in the literature, and thereby one of future directions is to pay more attention to this scenario.

{\em Nonconvex DOO and OG.} Nonconvex cost/constraint functions can be frequently encountered in applications (e.g., machine learning), although the current research mostly focuses on the convex case for DOO and OG. In this respect, a few works have studied the nonconvex case for centralized online learning. For example, locally Lipschitz and nonconvex cost functions were considered in \cite{lesage2020second} with $O(1+P_T)$ dynamic regret by proposing online Newton's method. Nonconvex cost functions, but satisfying local proximal-PL inequality (a generalization of Polyak-{\L}jasiewicz (PL) condition for unconstrained optimization), were studied in \cite{mulvaneydynamic} with $O(1+P_T)$ dynamic regret by developing online projected gradient descent with desirable initialization. However, only a few works have thus far investigated the nonconvex case in the distributed setup \cite{lu2019online,lu2021online,jiang2022distributed}. Consequently, it is nonnegligible to further take into account the nonconvex DOO and OG in future research directions.

{\em Adaptive Gradient Methods.} It is well known that adaptive methods are important in centralized optimization and deep learning due to its easy implementation and superior performance. The most popular adaptive gradient method is Adam by estimating first- and second-order moments of gradients \cite{kingma2014adam}. As a result, one natural idea is to apply adaptive methods to DOO and OG for improving the performance, which is exactly done in \cite{nazari2019adaptive,nazari2019dadam,carnevale2020distributed}. However, the research along this line is not far fully explored, leaving an enormous possibility for future directions.

{\em Second-order Methods.} It is easy to observe that all developed algorithms for DOO and OG are at most first-order methods, that is, depending on first-order gradients or zeroth-order function values. Nonetheless, it is well known that second-order methods, such as the Newton method, can generally improve the performance, usually outperforming first-order methods. As such, second-order online methods have been investigated in centralized online learning, e.g., \cite{hazan2007logarithmic,schraudolph2007stochastic,emiola2021sublinear,chang2021online,lesage2020second}. In contrast, the distributed case is yet to be explored.

{\em DOO on Riemannian Manifolds.} Aside from the Euclidean space studied for DOO in the literature, Riemannian manifolds, as a generalization of Euclidean spaces, have long been an intriguing topic in deep learning and centralized/distributed optimization. Riemannian manifolds possess a large number of applications such as in principal component analysis (PCA), independent component analysis (ICA), radar signal processing, dictionary learning, and mixture modeling \cite{smith1994optimization}. However, the study of DOO on Riemannian manifolds is still missing, only having a few works on the centralized setup \cite{maass2020online,wang2021no}, thereby posing the necessity of addressing this case in future.

{\em The Case with Switching Cost.} It is worth noting that it may incur a strategy changing cost when altering one's decision or action in practice, for example, the moving cost from the current position to the next selected position when the decision vector is the position of a robot. In this scenario, the local cost function of each agent $i\in[N]$ at time step $t$ is of the form
      \begin{align}
      f_{i,t}(x_{i,t},x_t^{-i})+d(x_{i,t},x_{i,t-1}),         \label{fd3}
      \end{align}
      where $x_{i,t}$ is still the decision vector of agent $i$ at time $t$, and $x_t^{-i}$ (different from $x_{-i,t}$) denotes a set of some other decision vectors except $x_{i,t}$. Note that $x_t^{-i}$ may be empty, or neighboring agents' decision vectors, or all other agents' decision vectors, such as the case of distributed online aggregative optimization \cite{li2021distributed2,carnevale2021distributed}. In (\ref{fd3}), $d(x_{i,t},x_{i,t-1})$ denotes a generic distance from $x_{i,t}$ to $x_{i,t-1}$, e.g., $\|x_{i,t}-x_{i,t-1}\|_1$, $\|x_{i,t}-x_{i,t-1}\|$, $\frac{1}{2}\|x_{i,t}-x_{i,t-1}\|^2$, and Bregman divergence $D_{\psi}(x_{i,t}|x_{i,t-1})$ for a strictly/strongly convex function $\psi$. Generally speaking, $f_{i,t}(x_{i,t},x_t^{-i})$ is called {\em hitting cost, operating cost}, or {\em stage cost} in the literature, and $d(x_{i,t},x_{i,t-1})$ is called {\em switching cost, smoothing cost}, or {\em movement cost} in the literature, which has been extensively studied in centralized online learning (e.g., \cite{goel2019beyond,li2020leveraging,shi2021competitive,wang2021online,zhao2021non}, to just name a few). However, it has thus far not been considered in DOO and OG, hence being regarded as one of interesting future research directions.

{\em The Case with Predictions.} In online learning, future information on cost functions is usually unaccessible and even adversarial. One natural question is if the performance can be improved when a few future information is available or can be predicted in some sense, including the gradient of next round at the current time and a lookahead window of future cost functions, etc. The answer is intuitively positive, usually leading to a lower constant improvement in big $O$ term, as confirmed in centralized online learning (e.g., \cite{lesage2020predictive,li2019online,li2020online}). In comparison, there are still no works on this study for DOO and OG until now, thereby motivating its possible investigation in future.

{\em Competitive Ratio.} It can be found that all related works in DOO and OG adapt the static and/or dynamic regrets as the performance metric. However, as presented in Section \ref{s2.1}, competitive ratio and adaptive regret are also considered as performance measures in the centralized online learning, although the relationships of all the metrics are yet to be fully understood, as discussed in the last paragraph of Section \ref{s2.1}. In this connection, it is interesting yet challenging to further address all the metrics (especially competitive ratio and adaptive regret) and their relationships in both centralized and distributed online learning for optimization and game.

{\em Online Game.} In contrast with DOO, online game in dynamic environments has so far been less explored in the literature, where local or private cost functions are time-varying. Currently, only a few works have researched online game with time-invariant and time-varying coupled inequality constraints in \cite{lu2020online,meng2021decentralized,meng2022decentralized}. Along this line, it is imperative to make more efforts to investigate online game in future, including improving regret bounds, studying individual static regret $S\text{-}Reg_i^g$ and $D\text{-}R_i^g,S\text{-}R_i^g$ as well as their relationships, and so forth.

{\em Multi-Agent Organization.} The organization of a multi-agent system usually means the collection of roles, relationships, and authority structures that are capable of governing its behavior, including hierarchies, holarchies, and teams organizations. The adaptivity to unpredictable environment changes by reorganizing towards the most appropriate organizations is also important \cite{horling2004survey,abbas2015organization}. In this respect, the agents' cooperative feature in DOO is similar to the team-based organization with the goal of achieving the entire team's mission \cite{horling2004survey}. And each agent's selfish behavior in OG is similar to the coalition-based organization with only one member in every coalition. However, to our best knowledge, the organizational design is not yet considered in DOO and OG, where the interactions among agents are simply described by stationary/time-varying graphs. Along this line, it is interesting to take the multi-agent organizational design into account in DOO and OG in future.


\section{Conclusion}

This paper has presented a comprehensive survey on DOO and OG, for which the overview has been performed from five viewpoints, that is, problem settings, communication issues, computation issues, algorithms, and up-to-date performances. They further include full state information, communication delays, asynchronous algorithms, privacy-preserving, security-guaranteeing, information quantization/compression, full gradient calculation, bandit feedback, projection-free algorithms, etc. With regard to these aspects, state-of-the-art results have been summarized and reported in this paper. Finally, possible future research directions have also been elaborated, which, hopefully, is conducive to further investigations on DOO and OG in future.

\section*{Acknowledgment}

The authors are grateful to the Editor, the Associate Editor and the anonymous reviewers for their insightful suggestions. The authors also would like to thank Mr. Yang Yu and Professor Min Meng for their assistance in numerical experiments.

\end{document}